\documentclass{tlp}
\usepackage{aopmath}
\usepackage{amsfonts}
\usepackage{epsf}
\usepackage{epic}
\usepackage{eepic}

\newtheorem{definition}{Definition}
\newtheorem{observation}{Observation}

\newcommand{\DefMath}{{\,\stackrel{\mbox{\rm\tiny def}}{=}\,}}

\newcommand{\Unit}{{\emptyset}}

\newcommand{\Consistent}{{\mbox{\em consistent\,}}}

\newcommand{\Holds}{{\mbox{\em holds\,}}}

\newcommand{\CalF}{{\cal F}}

\def\gothic{\gothicII
 \let\newumlaut=\gothicumlaut
 \def\ss{\kern-.1ex\char127\kern-.4ex z}
 \def\S{s}}

\font\gothicII=eufm10 scaled\magstep1

\newcommand{\ED}{\end{document}}

\newcommand{\comment}[1]{}

\newcommand{\Poss}{{\mbox{\rm\em Poss\/}}}
\newcommand{\PossF}{{\mbox{\footnotesize\rm\em Poss\/}}}

\newcommand{\Do}{{\mbox{\rm\em Do\/}}}

\newcommand{\True}{{\mbox{\rm\em True\/}}}
\newcommand{\False}{{\mbox{\rm\em False\/}}}

\newcommand{\Cancel}{{\mbox{\rm\em Cancel\/}}}
\newcommand{\Cancelled}{{\mbox{\rm\em Cancelled\/}}}
\newcommand{\Plus}{{\mbox{\rm\em Plus\/}}}
\newcommand{\Minus}{{\mbox{\rm\em Minus\/}}}
\newcommand{\OrHolds}{{\mbox{\rm\em OrHolds\/}}}
\newcommand{\Neq}{{\mbox{\rm\em Neq\/}}}
\newcommand{\NeqAll}{{\mbox{\rm\em NeqAll\/}}}

\newcommand{\OrNeq}{{\mbox{\rm\em OrNeq\/}}}
\newcommand{\Eq}{{\mbox{\rm\em Eq\/}}}

\renewcommand{\Holds}{{\mbox{\rm\em Holds\/}}}

\newcommand{\State}{{\mbox{\rm\em State\/}}}
\newcommand{\KState}{{\mbox{\rm\em KState\/}}}
\newcommand{\Kval}{{\mbox{\rm\em KnowsVal\/}}}
\newcommand{\KHolds}{{\mbox{\rm\em KHolds\/}}}

\renewcommand{\Consistent}{{\mbox{\rm\em Consistent\/}}}

\newcommand{\UNA}{{\mbox{\rm\em UNA\/}}}

\newcommand{\impl}{\supset}

\renewcommand{\iff}{\equiv}

\newcommand{\FluentSort}{{{\mbox{\sc fluent}}}}

\newcommand{\SitSort}{{{\mbox{\sc sit}}}}
\newcommand{\ActionSort}{{{\mbox{\sc action}}}}

\newcommand{\StateSort}{{{\mbox{\sc state}}}}

\newcommand{\NatSort}{{\mathbb{N}}}

\newcommand{\Fstate}{{\CalF_{\mbox{\scriptsize\rm\em state\/}}}}

\renewcommand{\impl}{\supset}
\renewcommand{\iff}{\equiv}

\renewcommand{\Holds}{{\mbox{\rm\em Holds\/}}}
\newcommand{\Knows}{{\mbox{\rm\em Knows\/}}}
\newcommand{\KnowsF}{{\mbox{\footnotesize\rm\em Knows\/}}}
\newcommand{\KnowsNot}{{\mbox{\rm\em KnowsNot\/}}}
\newcommand{\StateUpdate}{{\mbox{\rm\em StateUpdate\/}}}
\newcommand{\Perform}{{\mbox{\rm\em Perform\/}}}
\newcommand{\Execute}{{\mbox{\rm\em Execute\/}}}
\newcommand{\Update}{{\mbox{\rm\em Update\/}}}
\renewcommand{\State}{{\mbox{\rm\em State\/}}}

\renewcommand{\Fstate}{{\Sigma_{\mbox{\scriptsize\rm\em state\/}}}}

\newcommand{\LightPerception}{{\mbox{\rm\em Light\/}}}
\newcommand{\Light}{{\mbox{\rm\em Light\/}}}
\newcommand{\PiLight}{{\Pi_{\mbox{\rm\scriptsize\em Light\/}}}}
\newcommand{\Go}{{\mbox{\rm\em Go\/}}}
\newcommand{\At}{{\mbox{\rm\em At\/}}}
\newcommand{\Facing}{{\mbox{\rm\em Facing\/}}}
\newcommand{\Occupied}{{\mbox{\rm\em Occupied\/}}}
\newcommand{\Cleaned}{{\mbox{\rm\em Cleaned\/}}}
\newcommand{\Adjacent}{{\mbox{\rm\em Adjacent\/}}}
\newcommand{\SenseLoc}{{\mbox{\rm\em SenseLoc\/}}}
\newcommand{\Clean}{{\mbox{\rm\em Clean\/}}}
\newcommand{\Turn}{{\mbox{\rm\em Turn\/}}}

\newcommand{\GoInDirection}{{\mbox{\rm\em GoInDirection\/}}}
\newcommand{\Backtrack}{{\mbox{\rm\em Backtrack\/}}}
\newcommand{\TurnToGo}{{\mbox{\rm\em TurnToGo\/}}}
\newcommand{\Alter}{{\mbox{\rm\em Alter\/}}}
\newcommand{\Open}{{\mbox{\rm\em Open\/}}}

\newcommand{\Sd}{{\Sigma_{\mbox{\scriptsize\rm\em domain\/}}}}
\newcommand{\Pd}{\mbox{$P_{\mbox{\scriptsize\em domain}}$}}
\newcommand{\Ps}{\mbox{$P_{\mbox{\scriptsize\em strategy}}$}}
\newcommand{\Pk}{\mbox{$P_{\mbox{\scriptsize\em kernel}}$}}

\begin{document}
\bibliographystyle{acmtrans}

  \submitted{31 August 2002}
  \revised{7 November 2003, 23 April 2004}
  \accepted{9 August 2004}

\shorttitle

\title[FLUX: A Logic Programming Method for Reasoning Agents]
{FLUX: A Logic Programming Method for Reasoning Agents}
\author[M. Thielscher]
{MICHAEL THIELSCHER\\
Dresden University of Technology, 01062 Dresden, Germany \\
\email{mit@inf.tu-dresden.de}}
\pagerange{\pageref{firstpage}--\pageref{lastpage}}
\volume{\textbf{0} (0):}
\jdate{August 2002}
\setcounter{page}{1}
\pubyear{2002}

\maketitle
\label{firstpage}

\begin{abstract}
FLUX is a programming method for the design of agents that
reason logically about their actions and sensor information
in the presence of incomplete knowledge. The core of FLUX is
a system of Constraint Handling Rules, which enables agents
to maintain an internal model of their environment
by which they control their own behavior. The
general action representation formalism of the fluent calculus
provides the formal semantics for the constraint solver.
FLUX exhibits excellent computational behavior due to both a carefully
restricted expressiveness and the inference paradigm of progression.
\end{abstract}

\section{Introduction} \label{s:intro}

One of the most challenging and promising goals of Artificial Intelligence
research is the design of autonomous agents, including robots,
that explore partially known environments and that are able to act sensibly
under incomplete information. To attain this goal, the paradigm of Cognitive
Robotics~\cite{lesper:logica} is to endow agents with
the high-level cognitive capability of reasoning.
Exploring their environment, agents need to reason when they interpret sensor
information, memorize it, and draw inferences from combined sensor data.
Acting under incomplete information,
agents employ their reasoning facilities for selecting the right actions.
To this end, intelligent agents form
a mental model of their environment, which they constantly
update to reflect the changes they have effected and the
sensor information they have acquired.

Having
agents maintain an internal world model is necessary if we want them to
choose their actions not only on the basis of the current status
of their sensors but also on the basis of what they have previously
observed or
done. Moreover, the ability to reason about sensor information is necessary
if properties of the environment
can only be observed indirectly and require the agent to combine observations
made at different stages.

While standard programming languages such as Java do not provide general
reasoning facilities for agents, logic programming (LP) constitutes the ideal
paradigm for designing agents that are capable of
reasoning about their actions~\cite{shanah:solvin}. Examples of existing
LP-systems that have been developed from general action theories are
GOLOG~\cite{levesq:golog,reiter:logicBOOK}, based on
the situation calculus~\cite{mccart:situat}, or the robot
control language developed in~\cite{shanah:high-l}, based
on event calculus~\cite{kowals:logicb}. However, a disadvantage of both
these systems is that knowledge of the current state
is represented indirectly via the initial conditions
and the actions which the agent has performed up to now.
As a consequence, each time a condition is evaluated in an agent program
the entire history of actions is involved in the computation. This
requires ever increasing computational effort as the
agent proceeds, so that this concept does not scale up well to long-term agent
control.

Having an explicit state representation as a fundamental concept, the fluent
calculus~\cite{A:11} offers an alternative theory as the formal
underpinnings for a high-level agent programming method.
In this paper, we present the logic programming method FLUX
(for\/: {\em \underline{Flu}ent E\underline{x}ecutor\/})
for the design of intelligent agents that reason about their actions
using the fluent calculus. A
constraint logic program, FLUX
comprises a method for encoding
incomplete states along with a technique of updating these
states according to a declarative specification of the elementary actions
and sensing capabilities of an agent. Atomic state knowledge is encoded in
a list with a tail variable, which signifies the incompleteness of the
state. Negative and
disjunctive state knowledge is encoded by {\em constraints\/}. We present
a set of Constraint Handling Rules (CHRs)~\cite{fruehw:theory} for
combining and simplifying these constraints. In turn, these rules
reduce to standard finite domain constraints
when handling variable arguments of individual state components. Appealing to
their declarative interpretation, our CHRs are verified against the
foundational axioms of the fluent calculus.

With its powerful constraint solver,
the underlying FLUX kernel provides general reasoning facilities,
so that the agent programmer can focus on specifying the application
domain and designing the high-level behavior.
Allowing for
concise programs and supporting modularity, our method promises to be
eminently suitable for programming complex strategies for artificial agents.
Thanks to a restricted expressiveness and a sound but incomplete
inference engine, reasoning in FLUX is linear in the size of the
internal state representation. FLUX therefore exhibits excellent computational
behavior. Thanks to the progression principle, FLUX scales up
particularly well to long-term control.

The paper is organized as follows\/: In Section~\ref{s:fc}, we recapitulate
the basic notions and notations of the fluent calculus
as the underlying theory for an LP-based approach to reasoning about
actions. In Section~\ref{s:chr}, we present a set of
CHRs for constraints expressing negative and disjunctive state
knowledge. We prove their correctness wrt.\
the foundational axioms of the fluent calculus. 
In Section~\ref{s:update}, the constraint solver is embedded into a logic
program for reasoning about actions, which, too, is verified
against the underlying semantics of the fluent calculus.
In Section~\ref{s:knowledge}, we integrate state knowledge and sensing
into FLUX. An example of a FLUX agent
program is given in Section~\ref{s:cleanbot},
in which we also present the results of experiments showing
the computational merits of our approach. We conclude in Section~\ref{s:disc}.

The constraint solver, the general FLUX system, and
the example agent program
are available for download at our web site \,{\tt www.fluxagent.org}.

\section{Reasoning about states and sensor input with the fluent calculus} \label{s:fc}

\newcommand{\MAN}{\mbox{\begin{picture}(2,5)\put(1,4.25){\circle{1.5}}\put(1,3.5){\line(0,-1){1.5}}\put(1,2){\line(-1,-2){1}}\put(1,2){\line(1,-2){1}}\put(1,2.75){\line(3,1){1}}\put(1,2.75){\line(2,-1){1}}\end{picture}}}

\newcommand{\LIGHT}{\mbox{\begin{picture}(5,2)\put(2.5,0){\line(0,1){2}}\put(2,0.3){\line(-1,1){1.2}}\put(3,0.3){\line(1,1){1.2}}\put(1.5,0){\line(-1,0){1.5}}\put(3.5,0){\line(1,0){1.5}}\end{picture}}}

\begin{figure}
\figrule
\begin{center} \setlength{\unitlength}{1.1mm}
\begin{picture}(89,42)(0,1) \small
  \put(2,5){\framebox(35,35){}}
  \multiput(2,12)(0,7){4}{\line(1,0){2.5}}
  \multiput(6.5,19)(0,7){3}{\line(1,0){5}}
    \put(6.5,12){\line(1,0){2.5}}
  \multiput(13.5,19)(0,7){3}{\line(1,0){5}}
    \put(16,12){\line(1,0){2.5}}
    \put(20.5,12){\line(1,0){5}}
  \multiput(20.5,19)(0,7){3}{\line(1,0){2.5}}
    \put(27.5,12){\line(1,0){5}}
  \multiput(30,19)(0,7){3}{\line(1,0){2.5}}
  \multiput(34.5,12)(0,7){4}{\line(1,0){2.5}}

  \multiput(9,5)(7,0){4}{\line(0,1){2.5}}
  \multiput(9,9.5)(7,0){3}{\line(0,1){2.5}}
    \put(30,9.5){\line(0,1){5}}
  \multiput(9,19)(7,0){3}{\line(0,1){2.5}}
    \put(30,16.5){\line(0,1){5}}
  \multiput(9,23.5)(7,0){4}{\line(0,1){5}}
  \multiput(9,30.5)(7,0){3}{\line(0,1){2.5}}
    \put(30,30.5){\line(0,1){5}}
  \put(30,37.5){\line(0,1){2.5}}

  \put(5.5,3){\makebox(0,0){$1$}}
  \put(12.5,3){\makebox(0,0){$2$}}
  \put(19.5,3){\makebox(0,0){$3$}}
  \put(26.5,3){\makebox(0,0){$4$}}
  \put(33.5,3){\makebox(0,0){$5$}}

  \put(0,8.5){\makebox(0,0){$1$}}
  \put(0,15.5){\makebox(0,0){$2$}}
  \put(0,22.5){\makebox(0,0){$3$}}
  \put(0,29.5){\makebox(0,0){$4$}}
  \put(0,36.5){\makebox(0,0){$5$}}

  \put(5.5,29.5){\makebox(0,0){\MAN}}
  \put(19.5,8.5){\makebox(0,0){\MAN}}
  \put(19.5,22.5){\makebox(0,0){\MAN}}
  \put(33.5,22.5){\makebox(0,0){\MAN}}

  \put(4,6){\framebox(3,2.5){}}
    \put(4.5,6){\arc{1}{0}{3.14159}} \put(6.5,6){\arc{1}{0}{3.14159}}
  \put(5.5,8.5){\arc{2.5}{3.14159}{0}}

  \put(54,5){\framebox(35,35){}}
  \multiput(54,12)(0,7){4}{\line(1,0){2.5}}
  \multiput(58.5,19)(0,7){3}{\line(1,0){5}}
    \put(58.5,12){\line(1,0){2.5}}
  \multiput(65.5,19)(0,7){3}{\line(1,0){5}}
    \put(68,12){\line(1,0){2.5}}
    \put(72.5,12){\line(1,0){5}}
  \multiput(72.5,19)(0,7){3}{\line(1,0){2.5}}
    \put(79.5,12){\line(1,0){5}}
  \multiput(82,19)(0,7){3}{\line(1,0){2.5}}
  \multiput(86.5,12)(0,7){4}{\line(1,0){2.5}}

  \multiput(61,5)(7,0){4}{\line(0,1){2.5}}
  \multiput(61,9.5)(7,0){3}{\line(0,1){2.5}}
    \put(82,9.5){\line(0,1){5}}
  \multiput(61,19)(7,0){3}{\line(0,1){2.5}}
    \put(82,16.5){\line(0,1){5}}
  \multiput(61,23.5)(7,0){4}{\line(0,1){5}}
  \multiput(61,30.5)(7,0){3}{\line(0,1){2.5}}
    \put(82,30.5){\line(0,1){5}}
  \put(82,37.5){\line(0,1){2.5}}

  \put(57.5,3){\makebox(0,0){$1$}}
  \put(64.5,3){\makebox(0,0){$2$}}
  \put(71.5,3){\makebox(0,0){$3$}}
  \put(78.5,3){\makebox(0,0){$4$}}
  \put(85.5,3){\makebox(0,0){$5$}}

  \put(52,8.5){\makebox(0,0){$1$}}
  \put(52,15.5){\makebox(0,0){$2$}}
  \put(52,22.5){\makebox(0,0){$3$}}
  \put(52,29.5){\makebox(0,0){$4$}}
  \put(52,36.5){\makebox(0,0){$5$}}

  \put(57.5,22.5){\makebox(0,0){\LIGHT}}
  \put(57.5,36.5){\makebox(0,0){\LIGHT}}
  \put(64.5,29.5){\makebox(0,0){\LIGHT}}
  \put(64.5,8.5){\makebox(0,0){\LIGHT}}
  \put(64.5,22.5){\makebox(0,0){\LIGHT}}
  \put(71.5,15.5){\makebox(0,0){\LIGHT}}
  \put(71.5,29.5){\makebox(0,0){\LIGHT}}
  \put(78.5,8.5){\makebox(0,0){\LIGHT}}
  \put(78.5,22.5){\makebox(0,0){\LIGHT}}
  \put(85.5,15.5){\makebox(0,0){\LIGHT}}
  \put(85.5,29.5){\makebox(0,0){\LIGHT}}
\end{picture}
\end{center}
\caption{\label{f:crp}Layout of a sample office floor and a scenario in which
  four offices are occupied. In the right hand side, the locations are depicted
  in which the robot senses light.}
\figrule
\end{figure}
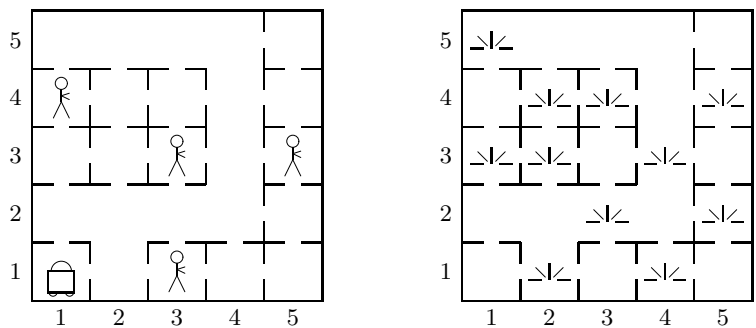

Throughout the paper, we will use
the following example of an agent in a dynamic environment\/:
Consider a cleaning robot which, in the evening, has to empty
the waste bins in the hallway and rooms of the floor of an office building.
The robot shall not, however, disturb anyone working in late. It is
equipped with a light sensor which is activated whenever the robot is adjacent
to a room that is occupied, without indicating which direction the
light comes from. An instance of this problem is depicted in
Figure~\ref{f:crp}. The robot can perform three basic actions, namely,
cleaning the current location, turning clockwise by~$90$~degrees, and
moving forward in the current direction to the adjacent cell.
Our task is to program the ``cleanbot''
to empty as many bins as possible without risking to burst into an
occupied office. This problem illustrates two challenges raised by incomplete
state knowledge\/: Agents have to act cautiously,
and they need to interpret and logically combine sensor information acquired
over time.

The fluent calculus is an axiomatic theory of actions
that provides the formal underpinnings for agents to
reason about their actions~\cite{A:11}. Formally, it is
a many-sorted predicate logic language which includes the two standard sorts of
a {\sc fluent} (i.e., an atomic state property) and a {\sc state}.
For the cleaning robot domain, for example,
we will use these four fluents (i.e., mappings into the
sort $\FluentSort\!$)\/:
$\At(x,y)\!$, representing that the robot is at~$(x,y)\!$;
$\Facing(d)\!$, representing that the robot faces
direction~$d\in\{1,\ldots,4\}$ (denoting, respectively, north, east, south, and
west); $\Cleaned(x,y)\!$, representing that the waste bin at~$(x,y)$ has been
emptied; and $\Occupied(x,y)\!$, representing that~$(x,y)$ is occupied.
We make the standard assumption of uniqueness-of-names,
$\UNA[\At,\Facing,\Cleaned,\Occupied]\!$.\footnote{Following
  \cite{baker:simple},
  $\!\UNA[h_1,\ldots,h_n]\,
  \DefMath\,\bigwedge_{i<j} h_i(\vec x)\not=h_j(\vec y)\,\wedge\,
  \bigwedge_{i}[h_i(\vec x)=h_i(\vec y)\impl \vec x=\vec y]\!$.}

States are built up from fluents (as atomic states) and their conjunction,
using the binary
function $\circ:\StateSort\times\StateSort\mapsto\StateSort$
along with the constant $\Unit:\StateSort$ denoting the empty state.
For example, the term $\At(1,1)\circ(\Facing(1)\circ z)$
represents a state in which the robot is in square $(1,1)$ facing north while
other fluents may hold, too, summarized in the variable
sub-state~$z\!$.\footnote{A word on the notation\/: Predicate and function
  symbols start with a capital letter while variables are denoted by lowercase
  letters, possibly with sub- or superscripts.
  Function~``$\!\circ\!$'' is written in infix notation.
  Throughout the paper, free variables in
  formulas are assumed universally quantified. Variables of sorts
  $\FluentSort$ and $\StateSort$
  shall be denoted, respectively, by the letters~$f$ and $z\!$.}

A fundamental notion is that of a fluent~$f$ to {\em hold\/} in
a state~$z\!$. For notational convenience,
the macro $\Holds(f,z)$ serves as an abbreviation for an equational
formula which says that~$z$ can be
decomposed into $f$ and some state~$z'$\/:
\begin{equation} \label{e:holds}
  \Holds(f,z)\,\DefMath\,(\exists z')\,z=f\circ z'
\end{equation}
This definition is accompanied by the following foundational
axioms of the fluent calculus, which ensure that
a state can be identified with the fluents that hold in it.
\begin{definition} \label{d:fstate}
  Assume a signature which includes the sorts $\FluentSort$ and
  $\StateSort$ such that $\FluentSort$ is a sub-sort of $\StateSort\!$, along
  with the functions $\circ,\Unit$ of sorts as above.
  The {\em foundational axioms $\Fstate$ of the fluent calculus\/}
  are\/:
  \begin{enumerate}
    \item Associativity and commutativity,
        \begin{equation} \label{e:ac}
        \begin{array}{rcl}
          (z_1\circ z_2)\circ z_3 & \!\!\!=\!\!\! & z_1\circ(z_2\circ z_3) \\
          z_1\circ z_2 & \!\!\!=\!\!\! & z_2\circ z_1
        \end{array}
        \end{equation}
    \item Empty state axiom,
        \begin{equation} \label{e:empty}
           \neg\Holds(f,\Unit)
        \end{equation}
    \item Irreducibility and decomposition,
        \begin{eqnarray}
          \Holds(f_1,f) & \!\!\!\impl\!\!\! & f_1=f \label{e:irred} \\
          \Holds(f,z_1\circ z_2) & \!\!\!\impl\!\!\! &
            \Holds(f,z_1)\vee\Holds(f,z_2) \label{e:decomp}
        \end{eqnarray}
    \item State equivalence and existence of states,
        \begin{eqnarray}
          (\forall f)\,(\Holds(f,z_1)\iff\Holds(f,z_2))\,\impl\,z_1=z_2 &&
             \label{e:stateequiv} \\
          (\forall P)(\exists z)(\forall f)\,(\Holds(f,z)\iff P(f))
        \hspace*{14.4mm}  &&
        \label{e:stateexistence}
        \end{eqnarray}
        where $P$ is a second-order predicate variable of
        sort $\FluentSort\!$.
  \end{enumerate}
\end{definition}
Axioms~(\ref{e:ac})--(\ref{e:decomp}) essentially characterize~``$\!\circ\!$''
as the union operation with~$\Unit$ as the empty set of fluents.
Associativity allows us to omit parentheses in nested applications
of~``$\!\circ\!$''. Axiom~(\ref{e:stateequiv}) says that two states
are equal if they contain the same fluents, and second-order
axiom~(\ref{e:stateexistence})
guarantees the existence of a state for any combination of
fluents.\footnote{A remark for readers who are familiar with
  early papers on the fluent calculus\/: The original solution to
  the frame problem in this calculus required function~``$\!\circ\!$''
  to be non-idempotent \cite{hoelld:deduct},
  so that, e.g., $\Occupied(2,3)\not=\Occupied(2,3)
  \circ\Occupied(2,3)\!$. Since
  this is against the intuition of~``$\!\circ\!$'' as
  a reified logical conjunction,
  the new axiomatization,
  first used in~\cite{A:15}, is no longer based on non-idempotence. In fact,
  foundational axiom~(\ref{e:stateequiv}) along with~(\ref{e:decomp})
  and associativity implies that $z\circ z=z$ for any~$z\!$.}

The foundational axioms of the fluent calculus
can be used to draw conclusions from incomplete state
specifications and acquired sensor information.
Consider, e.g., the definition of what it means for our cleaning robot to sense
light at a location $(x,y)$ in some state~$z\!$\/:
\begin{equation} \label{e:lipe} \begin{array}{l}
  \LightPerception(x,y,z)\,\iff \\
  \ \ \ \Holds(\Occupied(x+1,y),z)\vee\Holds(\Occupied(x,y+1),z) \\
  \ \ \ \vee\,\Holds(\Occupied(x-1,y),z)\vee\Holds(\Occupied(x,y-1),z)
\end{array} \end{equation}
Suppose that at the beginning the only given unoccupied locations are\/: the
home square of the robot (axiom~(\ref{e:11}) below), the
squares in the hallway (axiom~(\ref{e:alley}) below) and any location outside the
boundaries of the office floor (axioms~(\ref{e:boundx}),(\ref{e:boundy}) below).
Suppose further that the robot already went to clean~$(1,1)\!$, $(1,2)\!$,
and $(1,3)\!$, sensing light in the last square only (cf.\ Figure~\ref{f:crp}).
Thus the current state,~$\zeta\!$, satisfies
\begin{equation} \label{e:zeta}
  \!\!\!\zeta=\At(1,3)\circ\Facing(1)\circ\Cleaned(1,1)\circ\Cleaned(1,2)
    \circ\Cleaned(1,3)\circ z
\end{equation}
for some $z\!$, along with the following axioms\/:
\begin{eqnarray}
  && \!\!\!\!\neg\Holds(\Occupied(1,1),z) \label{e:11} \\
  && \!\!\!\!\neg\Holds(\Occupied(1,2),z)\,\wedge\,\ldots\,\wedge\,
     \neg\Holds(\Occupied(4,5),z)\ \ \ \ \ \ \label{e:alley} \\
  && \!\!\!\!(\forall x)\,(\neg\Holds(\Occupied(x,0),z)\wedge
     \neg\Holds(\Occupied(x,6),z)) \label{e:boundx} \\
  && \!\!\!\!(\forall y)\,(\neg\Holds(\Occupied(0,y),z)\wedge
     \neg\Holds(\Occupied(6,y),z)) \label{e:boundy} \\
  && \!\!\!\!\neg\LightPerception(1,2,\zeta)
      \label{e:nolight} \\
  && \!\!\!\!\LightPerception(1,3,\zeta) \label{e:light}
\end{eqnarray}
From (\ref{e:nolight}) and (\ref{e:lipe}) it follows
$\neg\Holds(\Occupied(1,3),\zeta)\!$.
With regard to~(\ref{e:zeta}), the
foundational axioms of decomposition~(\ref{e:decomp}) and
irreducibility~(\ref{e:irred}) along with the axiom of
uniqueness-of-names imply
\[
  \neg\Holds(\Occupied(1,3),z)
\]
On the other hand, (\ref{e:light}) and (\ref{e:lipe}) imply
\[ \begin{array}{l}
  \Holds(\Occupied(2,3),\zeta)\vee\Holds(\Occupied(1,4),\zeta) \\
  \vee\,\Holds(\Occupied(0,3),\zeta)\vee\Holds(\Occupied(1,2),\zeta)
\end{array} \]
Again with regard to~(\ref{e:zeta}),
the foundational axioms of decomposition and
irreducibility along with the axiom of uniqueness-of-names imply
\[ \begin{array}{l}
  \Holds(\Occupied(2,3),z)\vee\Holds(\Occupied(1,4),z) \\
  \vee\,\Holds(\Occupied(0,3),z)\vee\Holds(\Occupied(1,2),z)
\end{array} \]
From~(\ref{e:boundy}) and~(\ref{e:alley}) it follows that
\begin{equation} \label{e:occ}
  \Holds(\Occupied(2,3),z)\,\vee\,\Holds(\Occupied(1,4),z)
\end{equation}
This disjunction cannot be reduced further, that is, at this stage
the robot cannot decide whether the light in~$(1,3)$ comes from
office~$(2,3)$ or~$(1,4)$ (or both, for that matter).
Suppose, therefore, the cautious cleanbot goes back, turns east, and continues
with cleaning~$(2,2)\!$, which is a hallway location and therefore cannot be occupied 
according to~(\ref{e:alley}).
Sensing no light there (cf.\ Figure~\ref{f:crp}),
the new state is
\[ \begin{array}{l}
  \zeta'=\At(2,2)\circ\Facing(2)\,\circ \\
  \ \ \ \ \ \ \ \ \ \Cleaned(1,1)\circ\Cleaned(1,2)
  \circ\Cleaned(1,3)\circ\Cleaned(2,2)\circ z
\end{array} \]
for some $z$ that satisfies
(\ref{e:11})--(\ref{e:boundy}) and~(\ref{e:occ}). We also know that
$\neg\LightPerception(2,2,\zeta')\!$.
From~(\ref{e:lipe}), $\neg\Holds(\Occupied(2,3),\zeta')\!$;
hence, decomposition and
irreducibility along with the axiom of uniqueness-of-names imply
$\neg\Holds(\Occupied(2,3),z)\!$; hence, from~(\ref{e:occ}) it follows
$\Holds(\Occupied(1,4),z)\!$, that is,
now the robot can conclude that~$(1,4)$ is occupied.

\section{A constraint solver for the fluent calculus} \label{s:chr}

The axiomatic fluent calculus provides the formal underpinnings for
an LP-based approach to reasoning about incomplete state specifications.
To begin with, incomplete states are encoded by open-ended lists of fluents
(possibly containing variables)\/:
\begin{verbatim}
      Z = [F1,...,Fk | _ ]
\end{verbatim}
It is assumed that the arguments of fluents are encoded by integers
or symbolic constants,
which enables the use of a standard arithmetic solver for constraints
on partially known arguments.
Negative and disjunctive state knowledge is expressed by the
following {\em state constraints\/}\/:
\[  \begin{array}{l|l}
  \mbox{constraint} &
    \mbox{semantics} \\ \hline
  \verb:not_holds(F,Z): & \,\neg\Holds(f,z) \\
  \verb:not_holds_all(F,Z):\, & \,(\forall\vec x)\,\neg\Holds(f,z)\,,\
    \mbox{$\vec x$ variables in $f$} \\
  \verb:or_holds([F1,...,Fn],Z): & \,\bigvee_{i=1}^n\Holds(f_i,z) \\
\end{array} \]
These state constraints have been carefully designed so as to be
sufficiently expressive while allowing for efficient constraint solving.
An auxiliary constraint, written $\tt duplicate\_free(Z)\!$,
is used to stipulate that a list of fluents
contains no multiple occurrences. As an example, the following clause
encodes the specification of state~$\zeta$ of Section~\ref{s:fc} (cf.\
axioms~(\ref{e:zeta})--(\ref{e:light}))\/:

{\small
\begin{verbatim}
  zeta(Zeta) :-
    Zeta = [at(1,3),facing(1),cleaned(1,1),cleaned(1,2),cleaned(1,3) | Z],
    not_holds(occupied(1,1), Z),
    not_holds(occupied(1,2), Z), ..., not_holds(occupied(4,5), Z),
    not_holds_all(occupied(_,0), Z), not_holds_all(occupied(_,6), Z),
    not_holds_all(occupied(0,_), Z), not_holds_all(occupied(6,_), Z),
    duplicate_free(Zeta),
    light(1, 2, false, Zeta), light(1, 3, true, Zeta).
\end{verbatim}}
\noindent
The auxiliary predicate $\Light(x,y,p,z)$ defines under what circumstances
there is light ($\!p=\True\!$) or no light ($\!p=\False\!$)
in state~$z$ at square~$(x,y)$ (cf.\ axiom~(\ref{e:lipe})).

{\small
\begin{verbatim}
  light(X, Y, Percept, Z) :-
    XE #= X+1, XW #= X-1, YN #= Y+1, YS #= Y-1,
    ( Percept = false,
        not_holds(occupied(XE,Y), Z), not_holds(occupied(X,YN), Z),
        not_holds(occupied(XW,Y), Z), not_holds(occupied(X,YS), Z)
      ; Percept = true, or_holds([occupied(XE,Y),occupied(X,YN),
                                  occupied(XW,Y),occupied(X,YS)], Z) ).
\end{verbatim}}
\noindent
Here and in the following, we use the standard constraint language
of finite domains (see, e.g.,~\cite{henten:constr}),
which includes arithmetic constraints over the integers and symbolic constants,
using the equality, inequality, and ordering
predicates \,{\tt \#=}, {\tt\#\verb:\:=}, {\tt \#<}, {\tt \#>}\,
along with the arithmetic
functions \,{\tt +}, {\tt -}, {\tt *}; range constraints (written
\,{\tt X::}$\!a\!${\tt..}$\!b\!$); and logical combinations using
\,\verb:#/\:\,
and \,\verb:#\/:\, for conjunction and disjunction, respectively.
   
The state constraints are processed using
Constraint Handling Rules~\cite{fruehw:theory}. The general form of these rules
is
\[
    \verb:H1,...,Hm <=> G1,...,Gk | B1,...,Bn.:
\]
where
the {\em head\/} $H_1,\ldots,H_m$ is a sequence of constraints ($\!m\geq 1\!$);
the {\em guard\/} $G_1,\ldots,G_k$ is a sequence of Prolog literals
($\!k\geq 0\!$); and
the {\em body\/} $B_1,\ldots,B_n$ is a sequence of constraints
or Prolog literals ($\!n\geq 0\!$).
An empty guard is omitted; the empty body is denoted by $\True\!$.
The declarative interpretation of such a rule is given by the formula
\[
  (\forall\vec x)\,(G_1\wedge\ldots\wedge G_k\,\impl\,[
    H_1\wedge\ldots\wedge H_m\,\iff\,
    (\exists \vec y)\,(B_1\wedge\ldots\wedge B_n)]\,)
\]
where $\vec x$ are the variables in both guard and head and $\vec y$ are
the variables which additionally occur in the body.
The procedural interpretation of a CHR is given by a
transition in a constraint store\/: If the head can be matched against
elements of the constraint store and the guard can be derived,
then the constraints which match the head are replaced by the body.

\subsection{Handling negation}

\begin{figure}[t]
\figrule
\small
\begin{verbatim}
   not_holds(_,[])         <=> true.                                   %1
   not_holds(F,[F1|Z])     <=> neq(F,F1), not_holds(F,Z).              %2
   not_holds_all(_,[])     <=> true.                                   %3
   not_holds_all(F,[F1|Z]) <=> neq_all(F,F1), not_holds_all(F,Z).      %4

   not_holds_all(F,Z) \ not_holds(G,Z)     <=> instance(G,F) | true.   %5
   not_holds_all(F,Z) \ not_holds_all(G,Z) <=> instance(G,F) | true.   %6

   duplicate_free([])    <=> true.                                     %7
   duplicate_free([F|Z]) <=> not_holds(F,Z), duplicate_free(Z).        %8

   neq(F,F1)     :- or_neq(exists,F,F1).
   neq_all(F,F1) :- or_neq(forall,F,F1).

   or_neq(Q,Fx,Fy) :- Fx =.. [F|ArgX], Fy =.. [G|ArgY],
                      ( F=G -> or_neq(Q,ArgX,ArgY,D), call(D) ; true ).

   or_neq(_,[],[],(0#\=0)).
   or_neq(Q,[X|X1],[Y|Y1],D) :-
     or_neq(Q,X1,Y1,D1),
     ( Q=forall, var(X) -> ( binding(X,X1,Y1,YE) -> D=((Y#\=YE)#\/D1)
                                                  ; D=D1 )
                         ; D=((X#\=Y)#\/D1) ).

   binding(X,[X1|ArgX],[Y1|ArgY],Y) :- X==X1 -> Y=Y1
                                              ; binding(X,ArgX,ArgY,Y).
\end{verbatim}
\caption{\label{f:chr1}Constraint Handling Rules
  for the negation constraints and multiple occurrences of fluents.
  The notation \,{\tt H1}\,{\small$\!\setminus\!$}\,{\tt H2\,<=>\,G\,|\,B}
  is an abbreviation for
  {\,{\tt H1,H2\,<=>\,G\,|\,H1,B}}.}
\figrule
\end{figure}

Figure~\ref{f:chr1} depicts the first part of the constraint solver,
which contains the CHRs and auxiliary clauses for the two negation
constraints and the constraint on multiple occurrences.
In the following, these rules are proved correct wrt.\ the foundational
axioms of the fluent calculus.

To begin with, consider the auxiliary clauses, which define a
finite domain constraint that expresses the inequality of two fluent terms.
By $\OrNeq\!$, inequality of two fluents
with arguments ${\tt ArgX=[X1,\ldots,Xn]}$ and ${\tt ArgY=[Y1,\ldots,Yn]}$
is decomposed into the arithmetic constraint
${\tt X1\not=Y1}\vee\ldots\vee{\tt Xn\not=Yn}\!$.
Two cases are distinguished, depending on whether the variables in the first
term are existentially or universally quantified. In the latter case, a
simplified disjunction is generated, where the variables of the first fluent
are discarded while possibly giving rise to dependencies among the arguments
of the second fluent. Thus ${\tt neq\_all(f(\_,a,\_),f(U,V,W))}$ reduces to
${\tt a\not= V}\!$, and ${\tt neq\_all(f(X,X,X),f(U,V,W))}$ reduces to
${\tt U\not= V\vee V\not=W}\!$. To formally capture the universal quantification,
we define the notion of a {\em schematic\/} fluent $f=h(\vec x,\vec r)$ where
$\vec x$ denotes the variable arguments in~$f$
and $\vec r$ the non-variable arguments. The following observation
implies the correctness of the constraints generated by the auxiliary clauses.
\begin{observation} \label{o:neq}
  Consider a set $\CalF$ of functions into sort $\FluentSort\!$,
  a fluent $f_1=g(r_1,\ldots,r_m)\!$,
  a schematic fluent $f_2=g(x_1,\ldots,x_k,r_{k+1},\ldots,r_m)\!$,
  and a fluent $f=h(t_1,\ldots,t_n)\!$.
  Let $\Neq(f_1,f)\DefMath f_1\not=f$ and
  $\NeqAll(f_2,f)\DefMath(\forall x_1,\ldots,x_k)\,f_2\not=f\!$, then
  \begin{enumerate}
    \item if $g\not=h\!$, then $\UNA[\CalF]\,\models\,\Neq(f_1,f)$ and
        $\UNA[\CalF]\,\models\,\NeqAll(f_2,f)\!$;
    \item if $g=h\!$, then $m=n\!$, and $\UNA[\CalF]$ entails
          \[ \renewcommand{\arraystretch}{1.2} \begin{array}{rcl}
            \hspace*{-0.7cm}\Neq(f_1,f) & \!\!\!\! \iff \!\!\!\! &
            r_1\not=t_1\vee\ldots\vee r_m\not=t_n\vee 0\not=0 \\
            \hspace*{-0.7cm}\NeqAll(f_2,f) & \!\!\!\! \iff \!\!\!\! &
            [\bigvee_{\!\!\renewcommand{\arraystretch}{0.9}\tiny\begin{array}{c}\!\!i\!\not=\!j\!\!\\\!\!\!\!\!x_{i}\!=\!x_{j}\!\!\!\!\!\end{array}}t_i\not=t_j]\vee
            r_{k+1}\not=t_{k+1}\vee\ldots\vee r_m\not=t_n\vee 0\not=0
          \end{array} \]
  \end{enumerate}
\end{observation}
CHRs~1--4 in Figure~\ref{f:chr1}, by which negation constraints
are propagated, are then justified---on the basis of
their declarative interpretation---by
the foundational axioms of the fluent calculus.

\begin{proposition} \label{p:not_holds}
  Foundational axioms $\Fstate$ entail
  \begin{enumerate}
    \item $\!\neg\Holds(f,\Unit)\!$; and
    \item $\!\neg\Holds(f,f_1\circ z)\,\iff\,f\not=f_1\wedge\neg\Holds(f,z)\!$.
  \end{enumerate}
  Likewise, if $f=g(\vec x,\vec r)$ is a schematic fluent,
  then $\Fstate$ entails
  \begin{enumerate} \addtocounter{enumi}{2}
    \item $\!(\forall\vec x)\,\neg\Holds(f,\Unit)\!$; and
    \item $\!(\forall\vec x)\,\neg\Holds(f,f_1\circ z)\,\iff\,
        (\forall \vec x)\,f\not=f_1\,\wedge\,(\forall\vec x)\,
        \neg\Holds(f,z)\!$.
  \end{enumerate}
\end{proposition}
\begin{proof} Claim~1 follows by the empty state axiom. For claim~2
 we prove that $\Holds(f,f_1\circ z)$ iff $f=f_1\vee\Holds(f,z)\!$.
 The ``$\!\Rightarrow\!$'' direction
 follows by the
 foundational axioms of decomposition and irreducibility. For the
 ``$\!\Leftarrow\!$'' direction, suppose
 $f=f_1\!$, then $f_1\circ z=f\circ z\!$, hence
        $\Holds(f,f_1\circ z)\!$. Likewise, suppose $\Holds(f,z)\!$, then
        $z=f\circ z'$ for some~$z'\!$, hence $f_1\circ z=f_1\circ f\circ z'\!$,
        hence $\Holds(f,f_1\circ z)\!$.
  The proof of~3 and~4 is similar.
\end{proof}

CHRs 5 and~6, by which subsumed negative constraints are removed, are
correct since $(\forall\vec x)\,\neg\Holds(f_1,z)$ implies both
$\neg\Holds(f_2,z)$ and $(\forall\vec y)\,\neg\Holds(f_2,z)\!$, where
$f_1$ is a schematic fluent and $f_2$ is a fluent
such that $f_1\theta=f_2$ for some~$\theta\!$.
Finally, CHRs~7 and~8 for the auxiliary constraint on multiple occurrences
are correct since the empty list contains no duplicate elements and
a non-empty list contains no duplicates iff the head does not occur in the
tail and the tail itself is free of duplicates.

\subsection{Handling disjunction}

\begin{figure}[t]
\figrule
\small
\begin{verbatim}
  or_holds([F],Z) <=> F\=eq(_,_) | holds(F,Z).                         %9
  or_holds(V,Z)   <=> \+(member(F,V),F\=eq(_,_)) | or_and_eq(V,D),     %10
                                                   call(D).
               
  or_holds(V,[])  <=> member(F,V,W), F\=eq(_,_) | or_holds(W,[]).      %11

  or_holds(V,Z) <=> member(eq(X,Y),V),                                 %12
                    or_neq(exists,X,Y,D), \+ call(D) | true.
  or_holds(V,Z) <=> member(eq(X,Y),V,W),                               %13
                    \+ (and_eq(X,Y,D), call(D)) | or_holds(W,Z).

  not_holds(F,Z)     \ or_holds(V,Z) <=> member(G,V,W),                %14
                                         F==G          | or_holds(W,Z).
  not_holds_all(F,Z) \ or_holds(V,Z) <=> member(G,V,W),                %15
                                         instance(G,F) | or_holds(W,Z).

  or_holds(V,[F|Z])        <=> or_holds(V,[],[F|Z]).                   %16
  or_holds([F1|V],W,[F|Z]) <=> F1==F -> true ;                         %17
                               F1\=F -> or_holds(V,[F1|W],[F|Z]) ;
                               F1=..[_|ArgX], F=..[_|ArgY],
                                or_holds(V,[eq(ArgX,ArgY),F1|W],[F|Z]).
  or_holds([],W,[_|Z])     <=> or_holds(W,Z).                          %18

  and_eq([],[],(0#=0)).
  and_eq([X|X1],[Y|Y1],D) :- and_eq(X1,Y1,D1), D=((X#=Y)#/\D1).

  or_and_eq([],(0#\=0)).
  or_and_eq([eq(X,Y)|Eq],(D1#\/D2)) :- or_and_eq(Eq,D1), and_eq(X,Y,D2).

  member(X,[X|T],T).
  member(X,[H|T],[H|T1]) :- member(X,T,T1).
\end{verbatim}
\caption{\label{f:chr2}Constraint Handling Rules for the disjunctive constraint.}
\figrule
\end{figure}

Figure~\ref{f:chr2} depicts the second part of the constraint solver,
which contains the CHRs and auxiliary clauses for disjunctive state knowledge.
The solver employs an extended
notion of a disjunctive clause, where
each disjunction may include atoms of
the form $\Eq(\vec x,\vec y)$
in addition to fluents. The meaning of such a general disjunctive
constraint $\OrHolds([\delta_1,\ldots,\delta_k],z)$ is
\begin{equation} \label{e:or}
  \bigvee_{i=1}^k\left\{\begin{array}{ll}\Holds(f,z) & \mbox{if $\delta_i$
    is fluent $f$}\\\vec x=\vec y &
    \mbox{if $\delta_i$ is $\Eq(\vec x,\vec y)$}\end{array}\right.
\end{equation}
This generalization is needed for propagating disjunctions with variables
through compound states. Consider, as an example,
$\OrHolds([F(x),F(1)],[F(y)|z])\!$. This constraint will be rewritten to
$\OrHolds([\Eq([1],[y]),F(1),\Eq([x],[y]),F(x)],z)\!$,
in accordance with the fact that $\Fstate\cup\UNA[F]$ entails
\[ \begin{array}{c}
  \Holds(F(x),F(y)\circ z)\,\vee\,\Holds(F(1),F(y)\circ z) \\
  \iff \\
  x=y\,\vee\,\Holds(F(x),z)\,\vee\,1=y\,\vee\,\Holds(F(1),z)
\end{array} \]
which follows by the foundational axioms of irreducibility and decomposition.

CHR~9 in Figure~\ref{f:chr2} simplifies a singleton disjunction according
to~(\ref{e:or}). CHR~10 reduces a pure equational disjunction to a finite
domain constraint. Its correctness follows directly from~(\ref{e:or}), too.
CHR~11 simplifies a disjunction applied to the empty state.
It is justified by the empty state axiom, which
entails
\[
  [\Holds(f,\Unit)\vee\Psi]\,\iff\,\Psi
\]
for any formula~$\Psi\!$. CHRs~12 and~13 deal with disjunctions which include
an equality which is either true under any variable assignment,
or false. If the former, then the entire disjunction is true.
If, on the other hand,
the equality is necessarily false, then it is removed from the
disjunction. Correctness follows from
\[
  \vec x=\vec y\,\impl\,[(\vec x=\vec y\vee\Psi)\,\iff\,\top]
  \ \ \ \mbox{and}\ \ \ 
  \vec x\not=\vec y\,\impl\,[(\vec x=\vec y\vee\Psi)\,\iff\,\Psi]
\]
The next two CHRs are unit resolution steps\/: Rule~14
says that if a fluent~$f$ does not hold, then any disjunction that contains
an equal fluent~$g$ can be reduced by~$g\!$. Rule~15 generalizes this
to universally quantified negation constraints. The two CHRs are
justified, respectively, by
\[ \begin{array}{rcl}
  \neg\Holds(f,z) & \!\!\!\impl\!\!\! & [(\Holds(f,z)\vee\Psi)\,\iff\,\Psi] \\
  (\forall \vec x)\,\neg\Holds(f,z) & \!\!\!\impl\!\!\! &
    [(\Holds(g,z)\vee\Psi)\,\iff\,\Psi]\ \ \ \ \mbox{if}\ f\theta=g\
      \mbox{for some}\ \theta
\end{array} \]
where $\vec x$ are the variables in~$f\!$.

The last group of CHRs, 16--18, encode
the propagation of a disjunction
through a compound state. Informally speaking, each element in the disjunct
is compared to the head of the state and, if the two are unifiable, the
respective equational constraint is introduced into the disjunction.
Specifically,
with the help of the auxiliary ternary constraint
$\OrHolds(v,w,[f|z])\!$, a disjunction is divided into two parts.
List~$v$ contains the fluents that have not yet been
evaluated against the head~$f$ of the state.
List~$w$ contains those fluents that have been evaluated.
Thus the meaning of a ternary expression $\OrHolds(\Delta_1,\Delta_2,[f|z])$ is
\begin{equation} \label{e:or3}
  \OrHolds(\Delta_1,[f|z])\,\vee\,\OrHolds(\Delta_2,z)
\end{equation}
In the special case that disjunction~$\Delta_1$ contains a fluent~$f_1$ which
is identical to the head~$f$ of the state, disjunction~(\ref{e:or3})
is necessarily true and, hence, is resolved to $\True$ by CHR~17.
Otherwise, any fluent $f_1$ in~$\Delta_1$ which does not unify with~$f$
is propagated without inducing an equality. Any fluent $f_1$ which does unify
with $f$ extends the disjunction by the equality of the arguments of
$f_1$ and $f\!$. Recall, for example,
the constraint $\OrHolds([F(x),F(1)],[F(y)|z])$ mentioned earlier, which is
propagated thus\/:
\[ \renewcommand{\arraystretch}{1.3} \begin{array}{ll}
  & \OrHolds([F(x),F(1)],[F(y)|z]) \\
  \stackrel{\mbox{\scriptsize\tt\%16}}{\longrightarrow}\!\! &
  \OrHolds([F(x),F(1)],[\,],[F(y)|z]) \\
  \stackrel{\mbox{\scriptsize\tt\%17}}{\longrightarrow}\!\! &
  \OrHolds([F(1)],[\Eq([x],[y]),F(x)],[F(y)|z]) \\
  \stackrel{\mbox{\scriptsize\tt\%17}}{\longrightarrow}\!\! &
  \OrHolds([\,],[\Eq([1],[y]),F(1),\Eq([x],[y]),F(x)],[F(y)|z]) \\
  \stackrel{\mbox{\scriptsize\tt\%18}}{\longrightarrow}\!\! &
  \OrHolds([\Eq([1],[y]),F(1),\Eq([x],[y]),F(x)],z)
\end{array} \]

The three rules for propagating a disjunction
are justified by the following proposition, where
item~1 is for CHR~16, items~2--4 are for the three cases considered
in CHR~17, and item~5 is for CHR~18.
\begin{proposition}
  Consider a fluent calculus signature with a set $\CalF$ of functions
  into sort $\FluentSort\!$. Foundational axioms $\Fstate$
  and uniqueness-of-names $\UNA[\CalF]$ entail each of the following,
  where $\Psi_1$ is of the form $\OrHolds(\Delta,[f|z])$ and
  $\Psi_2$ is of the form $\OrHolds(\Delta,z)\!$\/:
  \begin{enumerate}
    \item $\!\Psi_1\,\iff\,[\Psi_1\vee\OrHolds([\,],z)]\!$;
    \item $\![\Holds(f,f\circ z)\vee\Psi_1]\vee\Psi_2\,\iff\,\top\!$;
    \item $\!f_1\not=f\,\impl\,(\,[\Holds(f_1,f\circ z)\vee\Psi_1]
           \vee\Psi_2\,\iff\,\Psi_1\vee[\Holds(f_1,z)\vee\Psi_2]\,)\!$;
    \item $\![\Holds(F(\vec x),F(\vec y)\circ z)\vee\Psi_1]
           \vee\Psi_2\,\iff\,\Psi_1\vee[\vec x=\vec y\,\vee
           \Holds(F(\vec x),z)\vee\Psi_2]\!$;
    \item $\![\OrHolds([\,],[f|z])\,\vee\,\Psi_2]\,\iff\,\Psi_2\!$.
  \end{enumerate}
\end{proposition}
\begin{proof}
  Claims~1 and~5 are obvious. Claim~2 follows by the definition of $\Holds\!$.
  Claims~3 and~4 follow from the foundational axioms
  of decomposition and irreducibility along with $\UNA[\CalF]\!$.
\end{proof}

\subsection{Using the constraint solver}

The constraint solver constitutes a system for automated reasoning about
incomplete states and sensor information. As an example, evaluating the
specification from the beginning of Section~\ref{s:chr} results in

\begin{verbatim}
 ?- zeta(Zeta).

 Zeta=[at(1,3),facing(1),cleaned(1,1),cleaned(1,2),cleaned(1,3) | Z]

 Constraints:
 or_holds([occupied(1,4),occupied(2,3)], Z)
 ...
\end{verbatim}
Light at~$(1,3)$ thus implies that $(1,4)$ or $(2,3)$ is occupied, but
it does not follow which of the two.
Adding the information that there is no light in~$(2,2)\!$,
the system is able to infer that $(1,4)$ must be occupied\/:
\begin{verbatim}
 ?- zeta(Zeta), light(2, 2, false, Zeta).

 Zeta=[at(1,3),facing(1),cleaned(1,1),cleaned(1,2),cleaned(1,3),
       occupied(1,4) | Z]

 Constraints:
 not_holds(occupied(2,3), Z)
 ...
\end{verbatim}

Although the
CHRs in the FLUX constraint system are correct, they may not enable
agents to draw all conclusions that follow logically from a state
specification if the underlying arithmetic solver trades
full inference capabilities
for efficiency. In standard implementations this is indeed the case, because
a conjunction or
a disjunction is simplified only if one of its equations or disequations
is either necessarily true or necessarily false. As a crucial
advantage of these concessions we have designed an efficient inference
system\/: The computational effort
of evaluating a new constraint is {\em linear\/} in the size of
the constraint store.

\section{Inferring state update in FLUX} \label{s:update}

In this section, we embed our constraint solver into a logic program
for reasoning about the effects of actions based on the fluent calculus.
Generalizing previous approaches~\cite{hoelld:deduct,bibel:deduct}, the
fluent
calculus provides a solution to the fundamental frame problem in the
presence of incomplete states~\cite{A:11}. The key is a
rigorously axiomatic characterization of
addition and removal of (finitely many) fluents from incompletely
specified states. The following inductive definition
introduces the macro equation $z_1-\vartheta^{-}=z_2$ with
the intended meaning that state~$z_2$ is state~$z_1$ minus the fluents in
the finite state~$\vartheta^{-}\!$\/:
\[ \renewcommand{\arraystretch}{1.2} \begin{array}{rcl}
  z_1-\Unit=z_2 & \!\!\!\DefMath\!\!\! & z_2=z_1 \\
  z_1-f=z_2 & \!\!\!\DefMath\!\!\! &
    (z_2=z_1\,\vee\,z_2\circ f=z_1)\,\wedge\,\neg\Holds(f,z_2) \\
  z_1-(f_1\circ f_2\circ\ldots\circ f_n)=z_2 & \!\!\!\DefMath\!\!\! &
        (\exists z)\,(z_1-f_1=z\,\wedge\,z-(f_2\circ\ldots\circ f_n)=z_2)
\end{array} \]
The crucial item is the second one, which
defines removal of a single fluent $f$ using a case distinction\/:
Either $z_1-f$ equals $z_1$ (which applies in case $\neg\Holds(f,z_1)\!$),
or $z_1-f$ plus $f$ equals $z_1\!$ (which applies in case $\Holds(f,z_1)\!$).

A further
macro $z_2=(z_1-\vartheta^{-})+\vartheta^{+}$ means that state~$z_2$ is
state~$z_1$ minus the fluents in $\vartheta^{-}$ plus the fluents in
$\vartheta^{+}\!$\/:
\begin{equation} \label{e:pm} \begin{array}{rcl}
  z_2=(z_1-\vartheta^{-})+\vartheta^{+} & \!\!\!\DefMath\!\!\! &
    (\exists z)\,(z_1-\vartheta^{-}=z\,\wedge\,z_2=z\circ\vartheta^{+})
\end{array} \end{equation}
where both $\vartheta^{+},\vartheta^{-}$ are finitely many $\FluentSort$ terms
connected by~``$\!\circ\!$''. 

\begin{figure}[t]
\figrule
\small
\begin{verbatim}
   holds(F,[F|_]).
   holds(F,Z) :- nonvar(Z), Z=[F1|Z1], F\==F1, holds(F,Z1).

   holds(F,[F|Z],Z).
   holds(F,Z,[F1|Zp]) :- nonvar(Z), Z=[F1|Z1], F\==F1, holds(F,Z1,Zp).

   minus(Z,[],Z).
   minus(Z,[F|Fs],Zp) :- ( \+ not_holds(F,Z) -> holds(F,Z,Z1) ;
                           \+ holds(F,Z)     -> Z1 = Z ;
                           cancel(F,Z,Z1), not_holds(F,Z1) ),
                         minus(Z1,Fs,Zp).
   plus(Z,[],Z).
   plus(Z,[F|Fs],Zp) :- ( \+ holds(F,Z)     -> Z1=[F|Z] ;
                          \+ not_holds(F,Z) -> Z1=Z ;
                          cancel(F,Z,Z2), Z1=[F|Z2], not_holds(F,Z2) ),
                        plus(Z1,Fs,Zp).

   update(Z1,ThetaP,ThetaN,Z2) :- minus(Z1,ThetaN,Z), plus(Z,ThetaP,Z2).
\end{verbatim}
\caption{\label{f:flux}The foundational clauses for reasoning about actions.
  Auxiliary predicate~$\Cancel$ is defined in
  Figure~\protect\ref{f:cancel}.}
\figrule
\end{figure}

Figure~\ref{f:flux} depicts a set of clauses which encode the solution to
the frame problem on the basis of the constraint solver for the fluent
calculus. The program culminates in the predicate
$\Update(z_1,\vartheta^{+},\vartheta^{-},z_2)\!$, by which an
incomplete state~$z_1$ is updated to~$z_2$
by positive and negative effects $\vartheta^{+}$ and $\vartheta^{-}\!$,
respectively, according to macro~(\ref{e:pm}).
The first two clauses in Figure~\ref{f:flux} encode macro~(\ref{e:holds}).
Correctness of this definition follows from the foundational axioms of
decomposition and irreducibility. The ternary $\Holds(f,z,z')$ means
$\Holds(f,z)\wedge z'=z-f\!$. The following proposition implies that the
corresponding clauses are correct wrt.\ the macro definition of fluent removal,
under the assumption that lists of fluents are free of duplicates.
\begin{proposition}
  Axioms $\Fstate\cup\{z=f_1\circ z_1\,\wedge\,\neg\Holds(f_1,z_1)\}$
  entail
  \[ \begin{array}{l}
    \Holds(f,z)\wedge z'=z-f\,\iff \\
    \ \ \ \ \ f=f_1\,\wedge\,z'=z_1 \\
    \ \ \ \ \ \vee\
    (\exists z'')\,(f\not=f_1\wedge\Holds(f,z_1)\wedge z''=z_1-f
      \wedge z'=f_1\circ z'')
  \end{array} \]
\end{proposition}
\begin{proof} We distinguish two cases.

  Suppose $f=f_1\!$, then $\Holds(f,z)$ since $z=f_1\circ z_1\!$.
  If $z'=z-f\!$, then $z'=(f_1\circ z_1)-f_1$
  since $z=f_1\circ z_1\!$; hence,
  $z'=z_1$ since $\neg\Holds(f_1,z_1)\!$. Conversely, if $z'=z_1\!$, then
  $z'=(f_1\circ z_1)-f_1=z-f\!$.

  Suppose $f\not=f_1\!$. If $\Holds(f,z)$ and $z'=z-f\!$, then
  $\Holds(f,z_1)$ and $z'=(f_1\circ z_1)-f\!$; hence, there is some~$z''$
  such that $z''=z_1-f$ and $z'=f_1\circ z''\!$. Conversely, if
  $\Holds(f,z_1)\wedge z''=z_1-f\wedge z'=f_1\circ z''\!$, then
  $\Holds(f,f_1\circ z_1)$
  and $z'=(f_1\circ z_1)-f\!$; hence, $\Holds(f,z)\wedge z'=z-f\!$.
\end{proof}

Removal and addition of finitely many fluents is defined recursively in Figure~\ref{f:flux}.
The recursive clause for $\Minus$ says that if $\neg\Holds(f,z)$ is
unsatisfiable (that is, $f$ is known to hold in~$z\!$), then
subtraction of~$f$ is given by the definition of the
ternary $\Holds$ predicate. Otherwise, if $\Holds(f,z)$ is unsatisfiable
(that is, $f$ is known to be false in~$z\!$), then $z-f$ equals $z\!$.
If, however, the status of the fluent is not entailed by the state
specification at hand for~$z$, then partial information
of~$f$ in $\Phi(z)$ may not transfer to the resulting state $z-f$ and, hence,
needs to be cancelled. Consider, for example, the partial state specification
\begin{equation} \label{e:incomplete_state1}
  \Holds(F(y),z)\ \wedge\ [\,\Holds(F(A),z)\vee\Holds(F(B),z)\,]
\end{equation}
This formula does not entail
$\Holds(F(A),z)$ nor $\neg\Holds(F(A),z)\!$. So what can be inferred about the
state $z-F(A)\!$\/? Macro expansion of ``$\!-\!$'' implies that
$\Fstate$ and $\{(\ref{e:incomplete_state1})\}\cup\{z_1=z-F(A)\}$ entail
$\neg\Holds(F(A),z_1)\!$. But it does not follow whether $F(y)$ holds in $z_1\!$, nor
whether $F(B)$ does, because
\[ \begin{array}{l}
    {[\,y=A\,\impl\,\neg\Holds(F(y),z_1)\,]}\ \wedge \\
  {[\,y\not=A\,\impl\,\Holds(F(y),z_1)\,]\ \wedge} \\
  {[\,\neg\Holds(F(B),z)\,\impl\,\neg\Holds(F(B),z_1)\,]}\ \wedge \\
  {[\,\Holds(F(B),z)\,\impl\,\Holds(F(B),z_1)\,]}
\end{array} \]

For this reason,
all partial information concerning~$f$ in the current state~$z$
is cancelled
in the clause for $\Minus$
prior to asserting that $f$ does not hold in the resulting state.
The definition of cancellation of a fluent~$f$ is given in
Figure~\ref{f:cancel} as an extension of our system of CHRs.
In the base case, all negative and disjunctive
state information affected by~$f$ is resolved via the
constraint $\Cancel(f,z)\!$. The latter in turn is resolved by the
auxiliary constraint $\Cancelled(f,z)\!$, indicating that $z$ contains
no (more) state knowledge which is affected by~$f\!$.
In the recursive clause for $\Cancel(f,z_1,z_2)\!$,
each atomic, positive state information that unifies with~$f$ is cancelled.

\begin{figure}[t]
\figrule
\small
\begin{verbatim}
    cancel(F,Z1,Z2) :-
       var(Z1) -> cancel(F,Z1), cancelled(F,Z1), Z2=Z1
       ;
       Z1=[G|Z], ( F\=G -> cancel(F,Z,Z3), Z2=[G|Z3]
                   ;
                   cancel(F,Z,Z2) ).

    cancel(F,Z) \ not_holds(G,Z)     <=>              \+ F\=G | true.
    cancel(F,Z) \ not_holds_all(G,Z) <=>              \+ F\=G | true.
    cancel(F,Z) \ or_holds(V,Z)      <=> member(G,V), \+ F\=G | true.

    cancel(F,Z), cancelled(F,Z) <=> true.
\end{verbatim}
\caption{\label{f:cancel}Auxiliary clauses and CHRs for cancelling partial
  information about a fluent.}
\figrule
\end{figure}

In a similar fashion, the recursive clause for $\Plus$ in Figure~\ref{f:flux}
says that if $\Holds(f,z)$ is unsatisfiable (that is, $f$ is known to
be false in~$z\!$), then $f$ is added to~$z\!$; otherwise,
if $\neg\Holds(f,z)$ is unsatisfiable
(that is, $f$ is known to hold in~$z\!$), then $z+f$ equals $z\!$.
If the status of the fluent is not entailed by the state
specification at hand for~$z$, then all partial information
about~$f$ in~$z$ is cancelled prior to adding $f$
to the state and asserting that $f$ does not hold in the tail.

The definitions for $\Minus$ and $\Plus$ imply that a fluent to be removed
or added does not hold or hold, respectively, in the
resulting state. Moreover, cancellation does not affect the
parts of the state specification which do not
unify with the fluent in question. Hence, these parts continue to hold in the
state resulting from the update. The correctness of this encoding of update
follows from the macros for ``$\!-\!$'' and ``$\!+\!$'', which imply that
a fluent holds in the
updated state just in case it either holds in the original state and is not
subtracted, or it is added.

\section{Reasoning about actions} \label{s:knowledge}

In this section, we extend our basic programming system so as to
enable agents to reason about what they know and to infer the results
of actions involving sensor information. Reasoning about knowledge
is necessary for agents with incomplete information,
as they need to select actions according to what they know of the state of
the environment. The formal concept of state knowledge also allows
to specify the effects of sensing actions, which, rather than affecting the
state itself, provide the agent with
more information about it.

\subsection{Knowledge and sensing in the fluent calculus} \label{ss:kfc}

Adopted from the situation calculus~\cite{mccart:situat},
the two standard sorts $\ActionSort$
and $\SitSort$ (i.e., situations) are used in the fluent calculus
to represent, respectively,
actions and sequences of actions. Action sequences are rooted in an
initial situation, usually denoted by the constant~$S_0\/:\SitSort\!$.
The pre-defined function $\Do\/:\ActionSort\times\SitSort\mapsto\SitSort$
maps an action and a situation into the situation after the
action. The function symbol
$\State\/:\SitSort\mapsto\StateSort$ is unique to the fluent calculus and
links the two key notions of a state and a situation\/:
$\State(s)$ denotes the state in situation~$s\!$.

Inspired by a model of knowledge in
the situation calculus~\cite{moore:formal,scherl:knowle},
the predicate $\KState:\,\SitSort\times\StateSort$
has been introduced in \cite{B:24}.
An instance $\KState(s,z)$ means that, according to the knowledge of the
agent, $z$ is a possible state in situation~$s\!$. As an example, recall
the initial state of our cleaning robot as depicted in Figure~\ref{f:crp}.
For the sake of argument, suppose that the robot is told it would perceive
light in~$(1,3)\!$. The initial knowledge of the cleanbot can then
be specified by the following axiom, which defines the {\em knowledge state\/}
in situation~$S_0\!$\/:
\begin{equation} \label{e:kstate0}
\begin{array}{ll}
  \multicolumn{2}{l}{\!\!\!\!\!\!\!\!(\forall z_0)\,(\,\KState(S_0,z_0)\,\iff} \\
  \ (\exists z)\,(\!\!\!\!\! &
    z_0=\At(1,1)\circ\Facing(1)\circ z\ \wedge \\
  & (\forall x,y)\,\neg\Holds(\At(x,y),z)\,\wedge\,
    (\forall d)\,\neg\Holds(\Facing(d),z)\ \wedge \\
  & \neg\Holds(\Occupied(1,1),z)\ \wedge \\
  & \neg\Holds(\Occupied(1,2),z)\,\wedge\,\ldots\,\wedge\,
     \neg\Holds(\Occupied(4,5),z)\ \wedge\\
  & (\forall x)\,(\neg\Holds(\Occupied(x,0),z)\wedge
     \neg\Holds(\Occupied(x,6),z))\ \wedge \\
  & (\forall y)\,(\neg\Holds(\Occupied(0,y),z)\wedge
     \neg\Holds(\Occupied(6,y),z))\ \wedge \\
  & \Light(1,3,z_0)\,)\,)
\end{array}
\end{equation}
That is to say, initially possible are all states in which the
robot is at a unique position, viz.~$(1,1)\!$, facing a unique direction,
viz.~north ($\!1\!$),
and neither~$(1,1)$ nor any square in the hallway or outside the
boundaries can be occupied. The possible states are further constrained
by the knowledge that there is light at~$(1,3)\!$.
On the other hand, the agent has no further
prior knowledge as to which offices are occupied or if any location is
cleaned.

A universal property of knowledge is that it is correct. To this end,
a simple foundational axiom stipulates that the actual state is always among
the possible ones\/:
\begin{definition} \label{d:fknows}
  The {\em foundational axioms of the fluent calculus for knowledge\/}
  are $\Fstate$ as in Definition~\ref{d:fstate} (cf.\ Section~\ref{s:fc})
  augmented by
  \[
    (\forall s)\,\KState(s,\State(s))
  \]
\end{definition}
Based on the notion of possible states,
a fluent is known to hold in a situation (or not to hold)
just in case it is true (false, respectively) in all possible
states in that situation\/:\footnote{For the sake of
  simplicity, we only consider knowledge
  of fluent literals in this paper; see \cite{B:24}
  for the generic extension to
  knowledge of formulas.}
\begin{equation} \label{e:knows} \begin{array}{rcl}
  \Knows(f,s) & \!\!\!\!\DefMath\!\!\!\! &
    (\forall z)\,(\KState(s,z)\impl\Holds(f,z)) \\
  \Knows(\neg f,s) & \!\!\!\!\DefMath\!\!\!\! &
    (\forall z)\,(\KState(s,z)\impl\neg\Holds(f,z)) \\
\end{array} \end{equation}
For example, the axiomatization of the initial knowledge,~(\ref{e:kstate0}),
entails that the cleanbot knows it is at~$(1,1)$ not facing east, that is,
\[
  \Fstate\cup\{(\ref{e:kstate0})\}\ \models\
    \Knows(\At(1,1),S_0)\wedge\Knows(\neg\Facing(2),S_0)
\]
On the other hand, the cleanbot does not know that office~$(1,4)$ is occupied\/:
\[
  \Fstate\cup\{(\ref{e:kstate0})\}\ \models\
    \neg\Knows(\Occupied(1,4),S_0)
\]
This is so because there is a possible state~$z_0$ which satisfies the
right hand side of the equivalence in~(\ref{e:kstate0}) and in which
$\Occupied(1,4)$ does not hold.

A supplementary macro defines knowledge of a value of a fluent. An agent
has this knowledge
just in case a particular instance of the fluent in question is known\/:
\begin{equation} \label{e:kval}
  \Kval(\vec x,f,s)\ \DefMath\
    (\exists\vec x)\,\Knows((\exists \vec x_1)\,f,s)
\end{equation}
where $\vec x_1$ are the variables in~$f$ besides~$\vec x\!$, and
$\Knows((\exists \vec x_1)\,f,s)$ stands for the formula
$(\forall z)\,(\KState(s,z)\impl
(\exists \vec x_1)\,\Holds(f,z))\!$.
For example, the axiomatization of the initial knowledge
entails that the cleanbot knows which direction it faces,
\[
  \Fstate\cup\{(\ref{e:kstate0})\}\ \models\
    \Kval(d,\Facing(d),S_0)
\]
On the other hand, although it knows that some office must be occupied, i.e.,
\[
  \Fstate\cup\{(\ref{e:kstate0})\}\ \models\
    \Knows((\exists x,y)\,\Occupied(x,y),S_0)
\]
the cleanbot does not know which one,
\[
  \Fstate\cup\{(\ref{e:kstate0})\}\ \models\
    \neg\Kval((x,y),\Occupied(x,y),S_0)
\]
This is so because there exists a possible state~$z_0$ which satisfies the
right hand side of the equivalence in~(\ref{e:kstate0}) and in which
$\Occupied(1,4)$ is the only positive instance of this fluent; and there also
exists a possible state in which a different one, viz.\
$\Occupied(2,3)\!$, is the only positive instance of this fluent.

While the definitions of knowledge by macros~(\ref{e:knows}) and~(\ref{e:kval})
are similar to the approach in the situation calculus~\cite{scherl:knowle}, a
crucial difference is that the latter defines knowledge in terms of possible
{\em situations\/}. To this end, the binary relation $K(s,s')$ is used with the
intuitive meaning that as far as the agent knows in situation~$s\!$,
it could as well be in situation~$s'\!$. This allows for a nested definition
of $\Knows\!$, which provides a form of introspection that is not supported
in the fluent calculus. On the other hand, the full expressiveness of modal
logic is computationally demanding. The notion of possible states allows for
a straightforward and---based on the results of the previous sections---tractable
implementation of knowledge, which is crucial for practical purposes.
We refer to~\cite{B:24}
for a more detailed comparison between the two approaches.

\subsection{Inferring knowledge in FLUX} \label{ss:FLUXknows}

The concept of knowing properties of the state is essential for the
evaluation of conditions in
agent programs under incomplete information.
By definition, a property is known just in case it is true in
all possible states. From a computational perspective, it is of course
impractical to evaluate a condition by literally checking every possible state,
since there is usually quite a number, often even infinitely many of them.
Fortunately, our constraint solver provides a
feasible alternative. Instead of verifying that
all states satisfy a property, we can just as well prove that the
{\em negation\/} of the property is {\em unsatisfiable\/} under a given
knowledge state. This suggests an elegant way of
encoding knowledge in FLUX using the principle of negation-as-failure.
To begin with, a knowledge state
$\KState(\sigma,z)\iff\Phi(z)$ is identified with the (incomplete) state
specification $\Phi(z)\!$. Then a fluent~$f$ is known in situation~$\sigma$
iff
the axiom set $\{\Phi(z),\neg\Holds(f,z)\}$
is unsatisfiable. Likewise,
$f$ is known to be false in situation~$\sigma$ iff
$\{\Phi(z),\Holds(f,z)\}$ is unsatisfiable.
\begin{theorem} \label{t:knowledge}
  Let $\KState(\sigma,z)\iff\Phi(z)$ be a knowledge state and $f$ a
  fluent, then
  \[
    \{\KState(\sigma,z)\iff\Phi(z)\}\,\models\,\Knows(f,\sigma)\ \ \mbox{iff}\ \
    \{\Phi(z),\neg\Holds(f,z)\}\,\models\,\bot
  \]
  and
  \[
    \{\KState(\sigma,z)\iff\Phi(z)\}\,\models\,\Knows(\neg f,\sigma)\ \ \mbox{iff}\ \
    \{\Phi(z),\Holds(f,z)\}\,\models\,\bot
  \]
\end{theorem}
\begin{proof}
  \[ \renewcommand{\arraystretch}{1.2} \begin{array}{ll}
    & \{\KState(\sigma,z)\iff\Phi(z)\}\,\models\,\Knows(f,\sigma) \\
    \mbox{iff} & \{\KState(\sigma,z)\iff\Phi(z)\}\,\models\,(\forall z)\,(
      \KState(\sigma,z)\impl\Holds(f,z)) \\
    \mbox{iff} & \models\,(\forall z)\,(\Phi(z)\impl\Holds(f,z)) \\
    \mbox{iff} & \models\,\neg(\exists z)\,(\Phi(z)\wedge\neg\Holds(f,z)) \\
    \mbox{iff} & \{\Phi(z),\neg\Holds(f,z)\}\,\models\,\bot
  \end{array} \]
  The proof of the second part is similar.
\end{proof}

\begin{figure}[t]
\figrule
\begin{center}
{\small\begin{verbatim}
       knows(F, Z) :- \+ not_holds(F, Z).

       knows_not(F, Z) :- \+ holds(F, Z).

       knows_val(X, F, Z) :- k_holds(F, Z), \+ nonground(X).

       k_holds(F, Z) :- nonvar(Z), Z = [F1|Z1],
                        ( instance(F1, F), F = F1 ; k_holds(F, Z1) ).

\end{verbatim}}
\end{center}
\caption{\label{f:knowledge}Knowledge in FLUX.}
\figrule
\end{figure}

This result is a formal justification of concluding knowledge of~$f$
if the constraint solver derives
an inconsistency upon asserting state constraint $\neg\Holds(f,z)$ under state
specification $\Phi(z)\!$. Figure~\ref{f:knowledge} shows how this is realized
in FLUX by clauses for $\Knows(f,s)$ and
$\Knows(\neg f,s)$ as well as for knowing a value of
a fluent. More complex knowledge expressions, such as disjunctive
knowledge, can be defined and encoded in a similar fashion.
The clausal definition of $\Kval(\vec x,f,z)$
uses the auxiliary predicate
$\KHolds(f,z)\!$, which {\em matches\/} the fluent expression~$f$ against all
fluents that positively occur in state~$z\!$. If so doing grounds all
variables in~$\vec x\!$, then a value for these variables is known.

Recall, for example, the FLUX state specification at the beginning
of Section~\ref{s:chr}, encoding state
specification~(\ref{e:zeta})--(\ref{e:light}).
We can use FLUX to show that the robot knows that
  room~$(1,3)$ is not occupied, while it does not know
  that office~$(1,4)$ is free, nor that it is not so\/:
\begin{verbatim}
   ?- zeta(Zeta),
      knows_not(occupied(1,3), Zeta),
      \+ knows(occupied(1,4), Zeta), 
      \+ knows_not(occupied(1,4), Zeta).

   yes.
\end{verbatim}
  As an example for the FLUX definition of knowing a value, consider
  this incomplete state specification\/:
\begin{verbatim}
   init(Z0) :-
      Z0=[at(X,2),facing(2)|Z], X#=1 #\/ X#=2, duplicate_free(Z0).
\end{verbatim}
The corresponding axiom in fluent calculus is
\[ \KState(S_0,z_0)\iff(\exists x,z)\,(z_0=\At(x,2)\circ\Facing(2)\circ z\,
\wedge\,[x=1\vee x=2]) \]
It follows that $\Kval(d,\Facing(d),S_0)$ while
$\neg\Kval((x,y),\At(x,y),S_0)$ but
$\Kval(y,\At(x,y),S_0)\!$\/:
\begin{verbatim}
   ?- init(Z0),
      knows_val([D], facing(D), Z0),
      \+ knows_val([X,Y], at(X,Y), Z0),
      knows_val([Y], at(_,Y), Z0).

   D = 2
   Y = 2
\end{verbatim}

In theory, agents using the fluent calculus are logically omniscient.
Therefore, the general problem of inferring knowledge under incomplete states
is computationally
demanding, if not undecidable in the first-order case.
This is so because full theorem proving is required to this end.
The careful design
of the state constraints supported in FLUX and the incomplete constraint solver,
however, make the task computationally feasible. Since deciding
unsatisfiability of a set of
constraints is linear in the size of the constraint store,
inferring knowledge in FLUX is linear in the size of the state description.

\subsection{Knowledge update}

The frame problem for knowledge is solved in the fluent calculus by
axiomatizing the relation between the possible states before and after
an action~\cite{B:24}. The
effect of $A(\vec x)\!$, be it a sensing action or not,
on the knowledge of the agent is specified by a so-called {\em knowledge update
axiom\/},\footnote{Below, the standard predicate
  $\PossF:\,\ActionSort\times\StateSort$ denotes that an action
  is possible in a state. Macro $\KnowsF(\PossF(a),s)$ stands for the
  formula $(\forall z)\,(\KState(s,z)\impl\Poss(a,z))\!$.}
\begin{equation} \label{e:kua}
\begin{array}{l}
  \Knows(\Poss(A(\vec x)),s)\impl \\
  \ \ \ (\exists\vec y)(\forall z')\,[\KState(\Do(A(\vec x),s),z')\,\iff \\
  \ \ \ \ \ \ \ \ \ \ \ \ \ \ \ \ \ \ \ \
    (\exists z)(\KState(s,z)\wedge\Psi(z',z)\wedge\Pi(\vec y,z',\Do(A(\vec x),s)))\,]
\end{array} \end{equation}
where $\Psi$ specifies the physical state update while $\Pi$ restricts
the possible states so as to agree with the actual state $\State(\Do(A(\vec x),s))$
on the sensed properties and values~$\vec y\!$.

As an example, let the three actions of the cleaning robot be denoted by
  \[ \begin{array}{rll}
    \Clean\/:\!\! & \ActionSort &
      \mbox{empty waste bin at current location} \\
    \Turn\/:\!\! & \ActionSort &
      \mbox{turn clockwise by $90^\circ$} \\
    \Go\/:\!\! &
      \ActionSort & \mbox{move forward to adjacent square}
  \end{array} \]
  The action preconditions can be axiomatized as
  \begin{equation} \label{e:crpposs}
  \begin{array}{rll}
    \Poss(\Clean,z)\,\iff\!\!\! & \top \\
    \Poss(\Turn,z)\,\iff\!\!\! & \top \\
    \Poss(\Go,z)\,\iff\!\!\! & (\forall d,x,y)\,(\!\!\!\!\!
    & \Holds(\At(x,y),z)\wedge\Holds(\Facing(d),z) \\
    && \ \ \ \impl(\exists x',y')\,\Adjacent(x,y,d,x',y'))
  \end{array} \end{equation}
  in conjunction with the auxiliary axiom
  \begin{equation} \label{e:3:adjacent} \begin{array}{ll}
    \Adjacent(x,y,d,x',y')\,\iff\!\!\! &
      1\leq d\leq 4\,\wedge\,1\leq x,x',y,y'\leq 5\ \wedge \\
      & {[}\,d=1\,\wedge\,x'=x\,\wedge\,y'=y+1\ \vee \\
      & \ \,d=2\,\wedge\,x'=x+1\,\wedge\,y'=y\ \vee \\
      & \ \,d=3\,\wedge\,x'=x\,\wedge\,y'=y-1\ \vee \\
      & \ \,d=4\,\wedge\,x'=x-1\,\wedge\,y'=y\,{]}
  \end{array} \end{equation}
  That is to say, going forward requires the robot not to face the
  wall of the building while emptying a waste bin and making a
  quarter turn clockwise is always possible.

  The actions $\Clean$ and $\Turn$ of our cleanbot involve no sensing.
  The physical effects of these actions
  are specified by the following knowledge update axioms\/:
  \begin{equation} \label{e:crpkua1} \begin{array}{llll}
    \multicolumn{4}{l}{\Knows(\Poss(\Clean),s)\impl} \\
    \multicolumn{4}{l}{\ \ \ [\,\KState(\Do(\Clean,s),z')\iff} \\
    \ \ \ \ \ \ \ (\exists z)\,(\!\!\!\!\! &
      \multicolumn{3}{l}{\KState(s,z)\ \wedge} \\
    & \multicolumn{3}{l}{(\exists x,y)\,(\Holds(\At(x,y),z)\,\wedge\,
       z'=z+\Cleaned(x,y))\,)\,]} \\ \\
    \multicolumn{4}{l}{\Knows(\Poss(\Turn),s)\impl} \\
    \multicolumn{4}{l}{\ \ \ [\,\KState(\Do(\Turn,s),z')\iff} \\
    \ \ \ \ \ \ \ (\exists z)\,(\!\!\!\!\! &
      \multicolumn{3}{l}{\KState(s,z)\ \wedge} \\
    & (\exists d)\,(\!\!\!\!\! &
      \multicolumn{2}{l}{\!\Holds(\Facing(d),z)\ \wedge} \\
    && \multicolumn{2}{l}{\!z'=z-\Facing(d)+
      \Facing(d\,\mbox{\rm mod}\,4+1))\,)\,]}
  \end{array} \end{equation}
  Thus $z'$ is a possible state after cleaning or turning, respectively,
  just in case $z'$ is the
  result of cleaning or turning in one of the previously possible states~$z\!$.

  The following knowledge update axiom for~$\Go$ combines
  the physical effect of going forward with
  information about whether light is sensed at the new location\/:
  \begin{equation} \label{e:crpkua2} \begin{array}{lll}
    \multicolumn{3}{l}{\Knows(\Poss(\Go),s)\impl} \\
    \multicolumn{3}{l}{\ \ \ [\,\KState(\Do(\Go,s),z')\iff} \\
    \ \ \ \ \ \ \ (\exists z)\,(\!\!\!\!\! &
      \multicolumn{2}{l}{\KState(s,z)\ \wedge} \\
    & (\exists d,x,y,x',y')\,(\!\!\!\!\! &
      \Holds(\At(x,y),z)\,\wedge \\
    && \Holds(\Facing(d),z)\,\wedge \\
    && \Adjacent(x,y,d,x',y')\,\wedge \\
    && z'=z-\At(x,y)+\At(x',y'))\,)\ \wedge \\
    \multicolumn{3}{l}{\ \ \ \ \ \ \
      [\,\PiLight(z')\,\iff\,\PiLight(\State(\Do(\Go,s)))\,]\,]}
  \end{array} \end{equation}
  where the sensed property indicates whether or not
the robot perceives a light at its
  current location\/:
  \begin{equation} \label{e:pilight}
    \PiLight(z)\ \DefMath\ (\exists x,y)\,(\Holds(\At(x,y),z)\wedge\Light(x,y,z))
  \end{equation}
  Thus axiom~(\ref{e:crpkua2}) says that
  $z'$ is a possible state after going forward if $z'$
  is the result of doing this action in some previously possible state
  and there is light at the current location in $z'$ just in case
  it is so in the actual state $\State(\Do(\Go,s))\!$.

As an example of sensing a fluent value rather than a proposition,
consider the specification of a location sensor.
As a pure sensing action, self-location has no physical effect. In general, this
is indicated in a knowledge update axiom by the sub-formula
$(\exists z)\,(\KState(s,z)\wedge z'=z)$ describing the (empty) physical effect.
For the sake of
compactness, this sub-formula has been simplified to
$\KState(s,z')$ in the following axiom\/:
\begin{equation} \label{e:kualoc} \begin{array}{ll}
  \multicolumn{2}{l}{\Knows(\Poss(\SenseLoc),s)\impl} \\
  \ \ \ (\exists x,y)(\forall z')\,(\!\!\!\!\! &
    \KState(\Do(\SenseLoc,s),z')\iff \\
  & \ \ \ \KState(s,z')\wedge\Holds(\At(x,y),z')
\end{array} \end{equation}
Put in words, there exist coordinates $x,y$ such that the robot is at~$(x,y)$
in all possible states of the successor situation. (The foundational axiom
for knowledge of Definition~\ref{d:fknows} (Section~\ref{ss:kfc})
then implies that $(x,y)$ must also be the actual location of the robot.)

\subsection{Inferring knowledge update in FLUX}

Updating the knowledge state of a FLUX agent involves two steps,
the physical effect and the sensing result of an action.
Since knowledge states are identified with (incomplete) FLUX states as
discussed in Section~\ref{ss:FLUXknows},
knowledge update according to the
physical effect amounts to updating a FLUX state
specification in the way discussed in Section~\ref{s:update}.
Having inferred the physical effect of an action, agents need to
evaluate possible sensing results as part of the update.
To this end, the sensing outcome of an action is encoded by a (possibly empty)
list of individual {\em sensing results\/}. 
The result of sensing a proposition is either of the constants
$\True$ or $\False\!$. The result of sensing a value is a ground term of the
respective sort. For example, the sensing
result for knowledge update axiom~(\ref{e:crpkua2}) is encoded by
$[\pi]$ where $\pi\in\{\True,\False\}\!$,
depending on whether light is actually sensed at the new location.
The sensing result for knowledge update axiom~(\ref{e:kualoc}), on
the other hand, should be encoded by $[x,y]$ where $x,y:\NatSort\!$.

Based on the notion of sensing results,
knowledge update axioms are encoded in FLUX as definitions of the predicate
$\StateUpdate(z_1,A(\vec x),z_2,y)$
describing the update of state~$z_1$ to~$z_2$ according to
the physical effects of action~$A(\vec x)$ and
the sensing result~$y\!$.
As an example, Figure~\ref{f:crpsua}
depicts a FLUX encoding of the action precondition and
knowledge update axioms
for the cleaning robot domain. Neither $\Clean$ nor
$\Turn$ provides any sensor data. The sensing result for action $\Go$
is evaluated with the help of
the auxiliary predicate $\Light$ as defined in Section~\ref{s:chr}.
\begin{figure}
\figrule
\begin{center}
{\small\begin{verbatim}
             poss(clean, _).
             poss(turn, _).
             poss(go, Z) :-
                knows_val([X,Y], at(X,Y), Z),
                knows_val([D], facing(D), Z),
                adjacent(X, Y, D, _, _),

             state_update(Z1, clean, Z2, []) :-
                holds(at(X,Y), Z1),
                update(Z1, [cleaned(X,Y)], [], Z2).

             state_update(Z1, turn, Z2, []) :-
                holds(facing(D), Z1),
                (D#<4 #/\ D1#=D+1) #\/ (D#=4 #/\ D1#=1),
                update(Z1, [facing(D1)], [facing(D)], Z2).

             state_update(Z1, go, Z2, [Light]) :-
                holds(at(X,Y), Z1),
                holds(facing(D), Z1),
                adjacent(X, Y, D, X1, Y1),
                update(Z1, [at(X1,Y1)], [at(X,Y)], Z2),
                light(X1, Y1, Light, Z2).

             adjacent(X, Y, D, X1, Y1) :-
                [X,Y,X1,Y1] :: 1..5, D :: 1..4,
                (D#=1) #/\ (X1#=X) #/\ (Y1#=Y+1)      % north
                #\/
                (D#=2) #/\ (X1#=X+1) #/\ (Y1#=Y)      % east
                #\/
                (D#=3) #/\ (X1#=X) #/\ (Y1#=Y-1)      % south
                #\/
                (D#=4) #/\ (X1#=X-1) #/\ (Y1#=Y).     % west
\end{verbatim}}
\end{center}
\caption{\label{f:crpsua}FLUX encoding of the precondition and update axioms for the cleanbot.}
\figrule
\end{figure}

  Consider, for example, the initial FLUX state for the cleaning robot
  shown in Figure~\ref{f:init}. Suppose that when going north twice,
  the robot senses no light after the
  first action but after the second one. With the following query the cleanbot computes
  the knowledge update for this sequence of actions and the
  given sensing results\/:
\begin{figure}
\figrule
\begin{center}
{\small\begin{verbatim}
   init(Z0) :-
      Z0 = [at(1,1),facing(1) | Z],
      not_holds(occupied(1,1), Z),
      not_holds(occupied(2,1), Z),         % hallway
      ..., not_holds(occupied(4,5), Z),    %
      consistent(Z0).

   consistent(Z) :-
      holds(at(X,Y), Z, Z1), [X,Y] :: 1..5, not_holds_all(at(_,_), Z1),
      holds(facing(D), Z, Z2), [D] :: 1..4, not_holds_all(facing(_), Z2),
      not_holds_all(occupied(_,0), Z),
      not_holds_all(occupied(_,6), Z),
      not_holds_all(occupied(0,_), Z),
      not_holds_all(occupied(6,_), Z),
      duplicate_free(Z).
\end{verbatim}}
\end{center}
\caption{\label{f:init}Initial state specification for the cleanbot domain.
  The clause for $\Consistent(z)$ specifies general domain constraints, such
  as uniqueness of the robot's position and orientation.}
\figrule
\end{figure}
\begin{verbatim}
   ?- init(Z0), state_update(Z0, go, Z1, [false]),
                state_update(Z1, go, Z2, [true]).

   Z0 = [at(1,1),facing(1) | Z]
   Z1 = [at(1,2),facing(1) | Z]
   Z2 = [at(1,3),facing(1) | Z]

   Constraints:
   not_holds(occupied(1,3), Z)
   or_holds([occupied(2,3),occupied(1,4)], Z)
   ...
\end{verbatim}
  Thus the agent has evaluated the acquired sensor data and inferred
  its actual position according to the physical effects of $\Go\!$.

  As an example for inferring the update when sensing a value of a fluent,
  consider the following FLUX clause, which encodes knowledge update
  axiom~(\ref{e:kualoc}) for action $\SenseLoc\!$\/:
\begin{verbatim}
   state_update(Z, sense_loc, Z, [X,Y]) :- holds(at(X,Y), Z).
\end{verbatim}
  That is, no physical effect affects the state but the sensed value
  is incorporated into the specification. Suppose, for instance,
  the agent is uncertain as to whether it moved north or
  east from its initial location~$(1,1)\!$, while the subsequent position
  tracking reveals that it is at~$(1,2)\!$\/:
\begin{verbatim}
  init(Z0) :- Z0 = [at(1,1),facing(D) | _], D#=1 #\/ D#=2,
              consistent(Z0).

  ?- init(Z0), state_update(Z0, go, Z1, [false]),
               state_update(Z1, sense_loc, Z2, [1,2]).

  Z0 = [at(1,1),facing(1) | Z]
  Z1 = [at(1,2),facing(1) | Z]
  Z2 = [at(1,2),facing(1) | Z]

  Constraints:
  not_holds(occupied(1,3), Z)
  ...
\end{verbatim}
  Thus the agent has inferred its actual position and, hence, concluded that
  it is actually facing north. Incidentally, knowing the location also
  allows to infer that office~$(1,3)$ is not occupied, which follows
  from the observation that no light is sensed after the
  $\Go$ action.

\subsection{Defining knowledge update for actions with conditional effects}

FLUX agents rely on knowledge update axioms in order to maintain their internal
model of the environment. As this model is usually incomplete, the
update axioms need to be carefully encoded in FLUX so as to always lead to
a correct resulting knowledge state. In particular, when specifying
an action with conditional effects the programmer needs to define the
correct update for any possible knowledge the agent may have concerning
the fluents affected by the action.
Consider, for example, the action $\Alter(x)$ to alter the position of
a toggle switch. If~$x$ happens to be open (fluent $\Open(x)\!$), then
it will be closed afterwards (i.e., not $\Open\!$);
otherwise, i.e., if it is closed beforehand, then it will be
open after the action. Tacitly assuming that the action is always possible,
its conditional effect is specified in the fluent calculus by the following
knowledge update axiom\/:
\begin{equation} \label{e:alter} \begin{array}{ll}
  \multicolumn{2}{l}{\KState(\Do(\Alter(x),s),z')\,\iff} \\
  \ \ \ (\exists z)\,(\,\KState(s,z)\,\wedge\,[\!\!\!\!\! &
    \Holds(\Open(x),z)\wedge z'=z-\Open(x) \\
  & \vee \\
  & \neg\Holds(\Open(x),z)\wedge z'=z+\Open(x)\,]\,)
\end{array} \end{equation}

The FLUX encoding of this update axiom requires to distinguish
three kinds of knowledge states. In case the current knowledge entails
that switch~$x$ is open, the resulting knowledge state is obtained through
updating by negative effect $-\Open(x)\!$. Conversely, in case the
current knowledge entails that switch~$x$ is not open, the resulting
knowledge state is obtained through updating by positive effect $+\Open(x)\!$.
Finally, if the current knowledge state does not entail the status of the
switch, then this uncertainty transfers to the updated knowledge state.
Moreover, possible partial (e.g., disjunctive) information regarding the
position of the affected
switch is no longer valid and, hence, needs to be cancelled.
\begin{verbatim}
   state_update(Z1, alter(X), Z2, []) :-
      knows(open(X), Z1)     -> update(Z1, [], [open(X)], Z2) ;
      knows_not(open(X), Z1) -> update(Z1, [open(X)], [], Z2) ;
      cancel(open(X), Z1, Z2).
\end{verbatim}
For example,
\begin{verbatim}
   ?- not_holds(open(t1), Z0),
      or_holds([open(t2),open(t3)], Z0),
      state_update(Z0, alter(t1), Z1, []),
      state_update(Z1, alter(t2), Z2, []).

   Z2 = [open(t1) | Z0]

   Constraints:
   not_holds(open(t1), Z0)
\end{verbatim}
That is to say, while switch~$T_1$ is known to be open after altering its
position, it no longer follows, after altering~$T_2\!$, that $T_2$ or $T_3$ is
open.\footnote{Actually, the inferred knowledge state in this example
  is slightly weaker than what is implied by knowledge update
  axiom~(\ref{e:alter}). Suppose that initially~$T_2$ or~$T_3$ is open.
  Then it follows that after altering the position of~$T_2\!$, if
  $T_2$ is open then so is~$T_3\!$\/! This is so because if $T_2$
  is open after changing its position, it must have been closed initially, and
  hence $T_3$ was (and still is) open. The corresponding implication,
  i.e., $\Holds(\Open(T_2),z_2)\impl\Holds(\Open(T_3),z_2)\!$,
  is not entailed by the updated FLUX state. Fortunately, obtaining
  a weaker update specification---just like an incomplete inference
  engine---is not an obstacle towards
  sound agent programs. Since FLUX agents are controlled by what they
  know of the environment, a sound but incomplete knowledge state suffices
  to ensure that the agent draws correct conclusions. This is a consequence of
  the simple fact that everything that is known under a weaker knowledge
  state is also known under the stronger one.}

\section{A FLUX control program for the cleaning robot} \label{s:cleanbot}

In this section, we show how our LP-based approach to reasoning
about actions can be used
as the kernel for a high-level programming method
for the design of agents that reason about their actions.
These agents use the concept of a state as their mental
model of the world when controlling their own behavior.
As they move along, agents constantly update their world model
in order to reflect the changes they have effected and the sensor information
they have acquired.
Thanks to the extensive reasoning facilities provided by the kernel of FLUX
and in particular the constraint solver,
the language allows to implement complex strategies with concise and
modular programs.

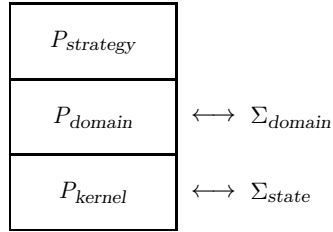
\begin{figure}
\figrule
\begin{center} \setlength{\unitlength}{0.2mm}
\begin{picture}(220,166)(0,20) 
  \put(0,120){\framebox(110,50){$\Ps$}}
    \put(135,95){\makebox(0,0){$\longleftrightarrow$}}
    \put(156,95){\makebox(0,0)[l]{$\Sd$}}
  \put(0,70){\framebox(110,50){$\Pd$}}
  \put(0,20){\framebox(110,50){{$\Pk$}}}
    \put(135,45){\makebox(0,0){$\longleftrightarrow$}}
    \put(156,45){\makebox(0,0)[l]{$\Fstate$}}
\end{picture}
\end{center}
\caption{\label{f:flux_architecture}The three components of FLUX agent programs.}
\figrule
\end{figure}

The general architecture of FLUX agent programs is depicted in
Figure~\ref{f:flux_architecture}. Every agent program contains the
kernel~$\Pk\!\!$, which consists of
\begin{itemize}
  \item the FLUX constraint system of Figure~\ref{f:chr1} and~\ref{f:chr2}
        plus a constraint solver for finite domains;
  \item the definition of update of Figure~\ref{f:flux}
        and~\ref{f:cancel};
  \item the definition of knowledge of Figure~\ref{f:knowledge}; and
  \item the following definition of execution,
        by which
        action~$a$ is performed and, simultaneously, the current state~$z_1$
        is updated to state~$z_2$ according to the effects and sensing
        result of performing~$a\!$\/:
        \begin{verbatim}
   execute(A, Z1, Z2) :-
      perform(A, Y), state_update(Z1, A, Z2, Y).\end{verbatim}
\end{itemize}

The second part,~$\Pd\!\!$, of a FLUX agent program contains encodings of
the domain axioms. These include
\begin{itemize}
  \item action precondition axioms,
  \item update axioms,
  \item domain constraints, and
  \item initial knowledge state.
\end{itemize}
The domain program for the cleanbot, for example, consists of the precondition
and update axioms of Figure~\ref{f:crpsua} along with the initial knowledge state
and domain constraints of Figure~\ref{f:init}.

On top of this, the programmer defines the intended
behavior of the agent via a control program~$\Ps\!\!$. This
program uses the basic predicate $\Execute(z_1,a,z_2)$ for
the execution of an action. To this end,
the interaction of the agent with the outside world needs to be
defined by the predicate $\Perform(a,y)\!$,
which causes the physical agent to carry out action~$a$ in the environment
such that~$y$ returns the sensing information acquired by performing this action.
Control programs~$\Ps$ use the predicate
$\Knows(f,z)$ (and its derivatives $\KnowsNot$ and $\Kval\!$)
to evaluate conditions against the internal world model.

Figure~\ref{f:cleanbot} depicts a sample control program for our cleaning robot.
After the initialization of the world model and
the execution of a $\Clean$ action at the home
square, the main loop is entered by which the robot systematically explores
and cleans the office floor. To this end, the program employs two
parameters containing, respectively, choice points yet to be explored
and the current path of the robot. The latter
is used to backtrack from a location
once all choices have been considered. A choice point is a list of
directions, which are encoded by $1$ (for north) to $4$ (for west) as usual. The path
is represented by the sequence, in reverse order, of the directions the
robot took in each step.

\begin{figure}
\figrule
\begin{center}
{\small\begin{verbatim}
      main :-
         init(Z0),
         execute(clean, Z0, Z1),
         Choicepoints = [[1,2,3,4]], Backtrack = [], 
         main_loop(Choicepoints, Backtrack, Z1).

      main_loop([Choices|Choicepoints], Backtrack, Z) :-
         Choices = [Direction|Directions] ->
         ( go_in_direction(Direction, Z, Z1)
           -> execute(clean, Z1, Z2),
              Choicepoints1 = [[1,2,3,4], Directions | Choicepoints],
              Backtrack1 = [Direction | Backtrack],
              main_loop(Choicepoints1, Backtrack1, Z2)
           ;
           main_loop([Directions|Choicepoints], Backtrack, Z) )
         ;
         backtrack(Choicepoints, Backtrack, Z).

      go_in_direction(D, Z1, Z2) :-
         knows_val([X,Y], at(X,Y), Z1),
         adjacent(X, Y, D, X1, Y1),
         \+ knows(cleaned(X1,Y1), Z1),
         knows_not(occupied(X1,Y1), Z1),
         turn_to_go(D, Z1, Z2).

      backtrack(_, [], _).
      backtrack(Choicepoints, [Direction|Backtrack], Z) :-
         Reverse is (Direction+1) mod 4 + 1,
         turn_to_go(Reverse, Z, Z1),
         main_loop(Choicepoints, Backtrack, Z1).

      turn_to_go(D, Z1, Z2) :-
         knows(facing(D), Z1) -> execute(go, Z1, Z2)
         ;
         execute(turn, Z1, Z), turn_to_go(D, Z, Z2).
\end{verbatim}}
\end{center}
\caption{\label{f:cleanbot}A cleanbot agent in FLUX.}
\figrule
\end{figure}

In the main loop, the cleanbot selects the first element of the current
choices. If the attempt to go into this direction is successful
(predicate $\GoInDirection\!$), then the robot empties the
waste bin at the new location. A new choice point is created,
and the backtrack path is augmented by the direction into which
the robot just went. If, on the
other hand, the chosen direction cannot be taken, then the main loop is
called with a reduced list of current choices. In case no more choices are
left, the cleanbot backtracks (predicate $\Backtrack\!$).

The auxiliary predicate $\GoInDirection(d,z_1,z_2)$ succeeds if the
cleanbot can safely go into direction~$d$ from its current location
in state~$z_1\!$, ending up in state~$z_2\!$. A direction is only explored
if the adjacent square is inside of the boundaries. Furthermore,
this location must not have
been visited already (that is, it is not known to be cleaned),
and---most importantly---the adjacent location must {\em known\/} not to
be occupied. By the auxiliary predicate $\Backtrack\!$, the robot takes back
one step on its current path by reversing the direction. The program terminates
once this path is empty, which implies that the robot has returned to its
home after it has visited and cleaned as many locations as possible.
The two auxiliary predicates $\GoInDirection$ and $\Backtrack$ in turn call
the predicate $\TurnToGo\!$, by which the robot makes turns
until it faces the intended direction, and then moves forward.

The following table illustrates what happens in the first nine calls to the main
loop when running the program with the initial state of Figure~\ref{f:init}
and the scenario depicted in Figure~\ref{f:crp}.
\begin{center}
\begin{tabular}{crrc}
  \multicolumn{1}{c}{$\At$} & 
    \multicolumn{1}{c}{Choicepoints} & \multicolumn{1}{c}{Backtrack}
    & \multicolumn{1}{c}{Actions} \\ \hline
  $(1,1)$ & {\tt [[1,2,3,4]]} & {\tt [\,]} & $GC$\\
  $(1,2)$ & {\tt [[1,2,3,4],[2,3,4]]} & {\tt [1]} & $GC$ \\
  $(1,3)$ & {\tt [[1,2,3,4],[2,3,4],[2,3,4]]} & {\tt [1,1]} & -- \\
  $(1,3)$ & {\tt [[2,3,4],[2,3,4],[2,3,4]]} & {\tt [1,1]} & -- \\
  $(1,3)$ & {\tt [[3,4],[2,3,4],[2,3,4]]} & {\tt [1,1]} & -- \\
  $(1,3)$ & {\tt [[4],[2,3,4],[2,3,4]]} & {\tt [1,1]} & -- \\
  $(1,3)$ & {\tt [[\,],[2,3,4],[2,3,4]]} & {\tt [1,1]} & $TTG$ \\
  $(1,2)$ & {\tt [[2,3,4],[2,3,4]]} & {\tt [1]} & $TTTGC$ \\
  $(2,2)$ & {\tt [[1,2,3,4],[3,4],[2,3,4]]} & {\tt [2,1]} & $TTTGC$
\end{tabular}
\end{center}

\vskip 1ex
\noindent
The letters $G,C,T$ are abbreviations for the actions $\Go\!$, $\Clean\!$,
and $\Turn\!$, respectively. After going north twice to office~$(1,3)\!$,
the cleanbot cannot continue in direction~$1$ or~$2$ because both
office~$(1,4)$ and office~$(2,3)$ may be occupied according to the robot's current
knowledge. Direction~$3$ is not explored since location~$(1,2)$ has
already been cleaned, and direction~$4$ is ruled out as $(0,3)$ is
outside of the boundaries. Hence, the cleanbot backtracks to~$(1,2)$ and
continues with the next choice there, direction~$2\!$,
which brings it to location~$(2,2)\!$. From there it goes north, and so on.
\begin{figure}[t]
\figrule
\begin{center} \setlength{\unitlength}{1.1mm}
\begin{picture}(40,40)(0,3) \small
  \put(2,5){\framebox(35,35){}}
  \multiput(2,12)(0,7){4}{\line(1,0){2.5}}
  \multiput(6.5,19)(0,7){3}{\line(1,0){5}}
    \put(6.5,12){\line(1,0){2.5}}
  \multiput(13.5,19)(0,7){3}{\line(1,0){5}}
    \put(16,12){\line(1,0){2.5}}
    \put(20.5,12){\line(1,0){5}}
  \multiput(20.5,19)(0,7){3}{\line(1,0){2.5}}
    \put(27.5,12){\line(1,0){5}}
  \multiput(30,19)(0,7){3}{\line(1,0){2.5}}
  \multiput(34.5,12)(0,7){4}{\line(1,0){2.5}}

  \multiput(9,5)(7,0){4}{\line(0,1){2.5}}
  \multiput(9,9.5)(7,0){3}{\line(0,1){2.5}}
    \put(30,9.5){\line(0,1){5}}
  \multiput(9,19)(7,0){3}{\line(0,1){2.5}}
    \put(30,16.5){\line(0,1){5}}
  \multiput(9,23.5)(7,0){4}{\line(0,1){5}}
  \multiput(9,30.5)(7,0){3}{\line(0,1){2.5}}
    \put(30,30.5){\line(0,1){5}}
  \put(30,37.5){\line(0,1){2.5}}

  \put(5.5,3){\makebox(0,0){$1$}}
  \put(12.5,3){\makebox(0,0){$2$}}
  \put(19.5,3){\makebox(0,0){$3$}}
  \put(26.5,3){\makebox(0,0){$4$}}
  \put(33.5,3){\makebox(0,0){$5$}}

  \put(0,8.5){\makebox(0,0){$1$}}
  \put(0,15.5){\makebox(0,0){$2$}}
  \put(0,22.5){\makebox(0,0){$3$}}
  \put(0,29.5){\makebox(0,0){$4$}}
  \put(0,36.5){\makebox(0,0){$5$}}

  \put(5.5,29.5){\makebox(0,0){\MAN}}
  \put(19.5,8.5){\makebox(0,0){\MAN}}
  \put(19.5,22.5){\makebox(0,0){\MAN}}
  \put(33.5,22.5){\makebox(0,0){\MAN}}

  \multiput(8,34.5)(7,0){5}{\circle{1}}
  \multiput(15,27.5)(7,0){4}{\circle{1}}
  \multiput(8,20.5)(7,0){2}{\circle{1}}
  \multiput(8,13.5)(7,0){5}{\circle{1}}
  \multiput(8,6.5)(7,0){2}{\circle{1}}
  \multiput(29,6.5)(0,7){3}{\circle{1}}

  \put(33.5,8.5){\makebox(0,0){\large\bf?}}

  \put(4,6){\framebox(3,2.5){}}
    \put(4.5,6){\arc{1}{0}{3.14159}} \put(6.5,6){\arc{1}{0}{3.14159}}
  \put(5.5,8.5){\arc{2.5}{3.14159}{0}}
  \put(5.5,9.5){\vector(0,-1){2.5}}
\end{picture}
\end{center}
\caption{\label{f:final_state}The final knowledge state in the
  cleaning robot scenario. The small circles indicate the cleaned locations.}
\figrule
\end{figure}
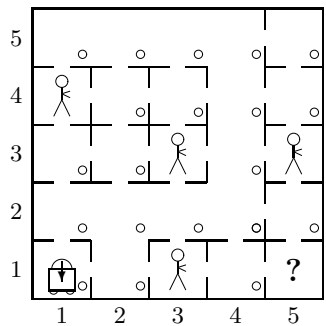
Figure~\ref{f:final_state} depicts the knowledge state at the time the
program terminates. Back home, the cleanbot has acquired knowledge
of all four occupied offices. Moreover, it has emptied all waste bins but
the ones in these four offices and the bin in office~$(5,1)\!$. This office
has not been visited because the robot cannot know that it is not
occupied---the light sensors have been activated at both surrounding
locations, $(4,1)$ and~$(5,2)\!$\/!

\subsection{Semantics of FLUX programs} \label{ss:semantics}

The semantics of a FLUX agent program is given as a combination of the fluent
calculus and the standard semantics of logic programming. We assume the reader
to be familiar with the basic notion of a computation tree for
constraint logic programs (see, e.g., \cite{jaffar:constr}).

Let $T$ be the computation tree for an agent program $\Ps\cup\Pd\cup\Pk$
along with a query $\{\leftarrow Q\}\!$. Tree~$T$ determines a
particular action sequence as follows. Let an {\em execution node\/}
be any node in~$T$ which starts with the atom~$\Execute\!$. Let
$\Execute(\alpha_1,\_,\_),\Execute(\alpha_2,\_,\_),\ldots$ be the ordered
sequence of all execution nodes occurring
in~$\!$T, then this tree is said to {\em generate\/}
the action sequence $\alpha_1,\alpha_2,\ldots$ This sequence is to be used when
proving formal properties of the agent program with the help of
the fluent calculus and the axiomatization~$\Sd$ of the application domain.
For example, a program can be called
{\em sound\/} if the domain axiomatization entails
that all actions are possible in the situation in which they are executed. Formally,
\[
  \Fstate\cup\Sd\,\models\,\Poss(\alpha_1,S_0)\wedge
    \Poss(\alpha_2,\Do(\alpha_1,S_0))\wedge\ldots
\]

Domain-dependent requirements are proved in a similar fashion.
The program for the cleanbot, for example, can be shown to admit a finite
computation tree; hence to terminate. Other properties are that
the cleanbot will always end up in its home~$(1,1)\!$, it will never enter an
office which is occupied (provided its light sensor functions correctly),
and it always cleans all locations in
the hallway. The formal proofs of these
properties are not deep but tedious, which is
why we refrain from giving them here.

\subsection{Computational Behavior} \label{ss:computational}

To illustrate the computational merits of FLUX, we have compared it
to GOLOG \cite{levesq:golog}, an agent programming language with similar
purposes. The cleanbot domain requires a variant of GOLOG which supports
incomplete states and sensing~\cite{reiter:knowle}. In this system,
incompletely specified initial situations are encoded by sets
of (propositional) prime implicates.
To decide whether a property is known to hold after
a sequence of actions, the property is {\em regressed\/}
to the initial situation.
If the resulting formula is entailed by the initial prime implicates,
then the original property is known to hold in the respective situation.
Acquired sensor information is regressed,
too, and the result is added to the initial set of prime implicates.

\begin{figure} \setlength{\unitlength}{1mm}
\figrule
\begin{center}
\begin{picture}(90,72)
    \put(50,36){\makebox(0,0){\epsfxsize=100mm\leavevmode\epsfbox{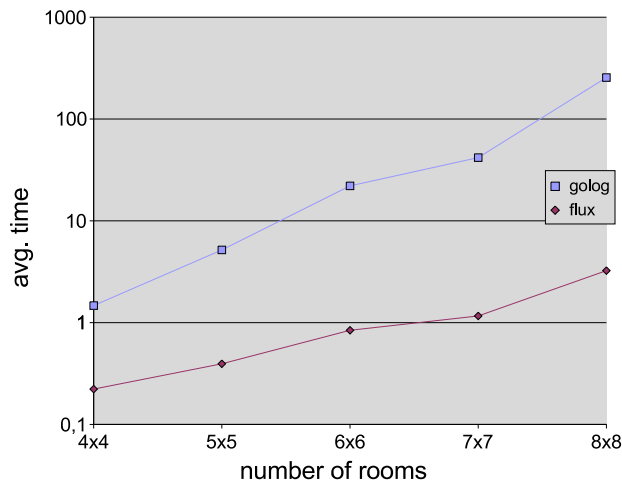}}}
\end{picture}
\end{center}
\caption{\label{f:fluxgolog1}Experimental results with the cleanbot control
program in FLUX
and GOLOG. (Notice the exponential scale on the vertical axis.)}
\figrule
\end{figure}

We have re-implemented the strategy of Figure~\ref{f:cleanbot}
for the cleanbot as a GOLOG program
and ran a series of
experiments with square office floors of different size.
For simplicity, no initial information about unoccupied
cells besides~(1,1) and the two adjacent ones were given to the robot.
Figure~\ref{f:fluxgolog1} depicts the results of five sets of experiments.
The given runtimes (seconds CPU time of a 1733~MHz processor)
are the average of
10~runs with randomly chosen occupied cells.

\begin{figure} \setlength{\unitlength}{1mm}
\figrule
\begin{center}
\begin{picture}(102,72)
    \put(51,36){\makebox(0,0){\epsfxsize=100mm\leavevmode\epsfbox{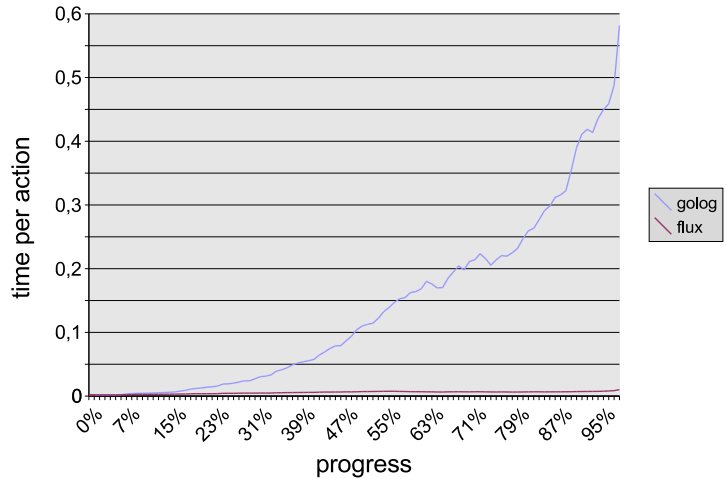}}}
\end{picture}
\end{center}
\caption{\label{f:fluxgolog2}Growth of the action selection
  time as the execution of the cleanbot program proceeds (averaged over 10~runs
  with $6\!\times\!6$ rooms).}
\figrule
\end{figure}

The dominance of FLUX has two main reasons\/:
\begin{enumerate}
  \item Since prime implicates can be used to encode arbitrary propositional
        formulas, the complexity of inferring knowledge in the
        GOLOG system of~\cite{reiter:knowle} is
        exponential. In contrast, the restricted first-order state
        representation and the incomplete inference engine of
        FLUX allows for inferring knowledge
        in linear time.
  \item In FLUX, the world model is {\em progressed\/} whenever
        an action is performed, and the new model is directly used to decide
        whether a property is currently known.
        The GOLOG system of~\cite{reiter:knowle}, on the other hand, is
        {\em regression-based\/}, so that deciding whether a property is known
        in a situation requires to regress the property through the
        previously performed actions. Consequently,
        the computational behavior of the GOLOG program worsens
        the longer the program runs. This can be clearly seen from the graphs in Figure~\ref{f:fluxgolog2},
        which depict the average time for action selection at different stages of the execution of
        the cleanbot program.
  \item To solve the frame problem, FLUX uses state update axioms,
        which specify the effects of an action on an entire state. When
        progressing a state through an update axiom, the large body of
        unaffected knowledge simply remains in the constraint store. This
        is what makes up an efficient solution to the frame problem even in the
        presence of incomplete states.
\end{enumerate}

\section{Discussion} \label{s:disc}

We have presented the logic programming method FLUX for the design of logically
reasoning agents. The agents use a system of Constraint Handling Rules and
finite domain constraints to reason about actions in the presence of incomplete
states. Both the constraint solver and
the logic program for state update have been formally verified
against the action theory of the fluent calculus. Thanks to a
carefully chosen expressiveness, the FLUX kernel exhibits excellent
computational behavior.

The closest related work is the programming language GOLOG~\cite{levesq:golog}
for dynamic domains,
which is based on the situation calculus and successor state axioms as a solution
to the frame problem~\cite{reiter:frame}. The main differences are\/:
\begin{enumerate}
  \item GOLOG defines a special programming language for strategies,
        while FLUX strategies are standard logic programs.
  \item With the exception of~\cite{reiter:knowle},
        existing implementations of GOLOG apply the principle of
        negation-as-failure to state specifications and, hence, are restricted
        to complete state knowledge and deterministic
        actions.
        With its underlying constraint solver, FLUX provides a natural
        way of representing and reasoning with incomplete states as well as
        nondeterministic actions.
  \item The logic programs for GOLOG described in the literature all apply
        the principle of regression to evaluate conditions in agent programs.
        While this is efficient for short action sequences,
        the computational effort increases with the number of performed actions.
        With the progression principle, FLUX programs
        scale up well to the control of
        agents over extended periods.\footnote{In order to achieve a similar
        behavior, GOLOG would have to be reimplemented by appealing to the
        definition of progression in the situation calculus
        of~\cite{lin:progre}, which, however, is not first-order
        definable in general.} Moreover, progression through state
        update axioms in FLUX provides an efficient solution to the
        frame problem, because unaffected state knowledge simply remains in the
        constraint store.
  \item GOLOG includes the concept of nondeterministic programs as a means
        to define a search space for a planning problem. To find a plan,
        such a program is executed ``off-line'' with the aim to find a
        run by which the planning goal is attained. A similar concept
        can be added to FLUX, allowing agents to interleave planning with
        program execution~\cite{B:30}.
\end{enumerate}

We are conducting experiments where FLUX is applied to the high-level control
of a real robot, whose task is to collect and deliver in-house mail in an
office floor~\cite{A:17}.
To this end, the logic programming system has been extended by
a solution to the qualification problem~\cite{mccart:episte} in
the fluent calculus which accounts for
unexpected failure of actions~\cite{A:15,B:28}. Future work will include the
gradual extension of the expressiveness
of FLUX, e.g., by constraints for exclusive
disjunction, without loosing the computational merits of the approach.

\section*{Acknowledgments}

The author wants to thank Stephan Schiffel for his help with the experiments
and Matthias Fichtner, Axel Gro\ss{}mann, Yves Martin, and the anonymous
reviewers for valuable comments on an earlier version.
Parts of the work reported in this paper have been carried out
while the author was a visiting researcher at the University of New South Wales
in Sydney, Australia.


\end{document}